\documentclass[review]{cas-sc}

\usepackage[numbers,sort&compress]{natbib}
\usepackage{subfigure}
\usepackage{caption}
\usepackage{amsmath}
\usepackage{hyperref}
\usepackage{stfloats}
\usepackage{wrapfig}
\usepackage{graphicx}
\DeclareGraphicsExtensions{.tif,.pdf,.png,.jpg}
\usepackage{cuted}
\usepackage{lipsum}
\usepackage{svg}
\usepackage{bbding}
\usepackage{setspace}
\usepackage{booktabs}
\usepackage{pifont}

\begin{document}
\let\WriteBookmarks\relax
\def\floatpagepagefraction{1}
\def\textpagefraction{.001}
\def\sssectionfont{\rmfamily\fontsize{10.5pt}{12pt}%
  \fontseries{b}\selectfont}
\let\printorcid\relax
\captionsetup[figure]{labelfont={bf},labelformat={default},labelsep=period,name={Fig.}}

\shorttitle{}

\shortauthors{Changsong Liu et~al.}

\title [mode = title]{Learning to utilize image second-order derivative information for crisp edge detection}                      



%
\author[1]{Changsong Liu}[style=chinese]

\author[1]{Yimeng Fan}[style=chinese]

\author[1]{Mingyang Li}[style=chinese]

\author[1]{Wei Zhang}[style=chinese]
\cormark[1]
\ead[]{tjuzhangwei@tju.edu.cn}

\author[2]{Yanyan Liu}[style=chinese]

\author[3]{Yuming Li}[style=chinese]

\author[1]{Wenlin Li}[style=chinese]

\author[4,5]{Liang Zhang}[style=chinese]






\affiliation[1]{organization={School of Microelectronics, Tianjin University},
    city={Tianjin},
    postcode={300072}, 
    country={China}}




\affiliation[2]{organization={College of Electronic Information and Optical Engineering, Nankai University},
    city={Tianjin},
    postcode={300072}, 
    country={China}}

\affiliation[3]{organization={College of Information Engineering, Luoyang Polytechnic},
    city={Luoyang},
    postcode={471000}, 
    country={China}}

\affiliation[4]{organization={Tianjin Fire Science and Technology Research Institute of MEM},
    city={Tianjin},
    postcode={300381},
    country={China}}

\affiliation[5]{organization={School of Transportation Science and Engineering, Beihang University},
    city={Beijing},
    postcode={100191},
    country={China}}



\cortext[cor1]{Corresponding author.}



\begin{abstract}
Edge detection is a fundamental task in computer vision. It has made great progress under the development of deep convolutional neural networks (DCNNs), some of which have achieved a beyond human-level performance. However, recent top-performing edge detection methods tend to generate thick and noisy edge lines. In this work, we solve this problem from two aspects: (1) the lack of prior knowledge regarding image edges, and (2) the issue of imbalanced pixel distribution. We propose a second-order derivative-based multi-scale contextual enhancement module (SDMCM) to help the model locate true edge pixels accurately by introducing the edge prior knowledge. We also construct a hybrid focal loss function (HFL) to alleviate the imbalanced distribution issue. In addition, we employ the conditionally parameterized convolution (CondConv) to develop a novel boundary refinement module (BRM), which can further refine the final output edge maps. In the end, we propose a U-shape network named LUS-Net which is based on the SDMCM and BRM for crisp edge detection. We perform extensive experiments on three standard benchmarks, and the experiment results illustrate that our method can predict crisp and clean edge maps and achieves state-of-the-art performance on the BSDS500 dataset (ODS=0.829), NYUD-V2 dataset (ODS=0.768), and BIPED dataset (ODS=0.903).
\end{abstract}




\maketitle

\section{Introduction}
\label{Introduction}
Edge detection is a fundamental problem in computer vision and image processing, playing a crucial role in a wide range of applications, from autonomous driving to medical imaging. It serves as the cornerstone for many high-level computer vision tasks by identifying the boundaries and structures within images, such as salient object detection \cite{yao2023object,zheng2023boundary}, object recognition \cite{feng2023boundary,fan2024sfod}, and image segmentation \cite{jin2023edge,xiao2023baseg}. With the rapid development of deep learning techniques, numerous researchers aim to employ Deep Convolutional Neural Networks (DCNNs) for edge detection. Recent studies have proposed several remarkable DCNN-based methods, such as HED \cite{xie2015holistically}, RCF \cite{liu2017richer}, and DexiNed \cite{soria2023dense}. These DCNN-based approaches have achieved state-of-the-art performance on a number of benchmarks including BSDS500 \cite{arbelaez2011contour} and NYUD-V2 \cite{silberman2012indoor}, which demonstrate the strong power of DCNN.

\begin{figure}
    \centering
    \includegraphics[width=\textwidth]{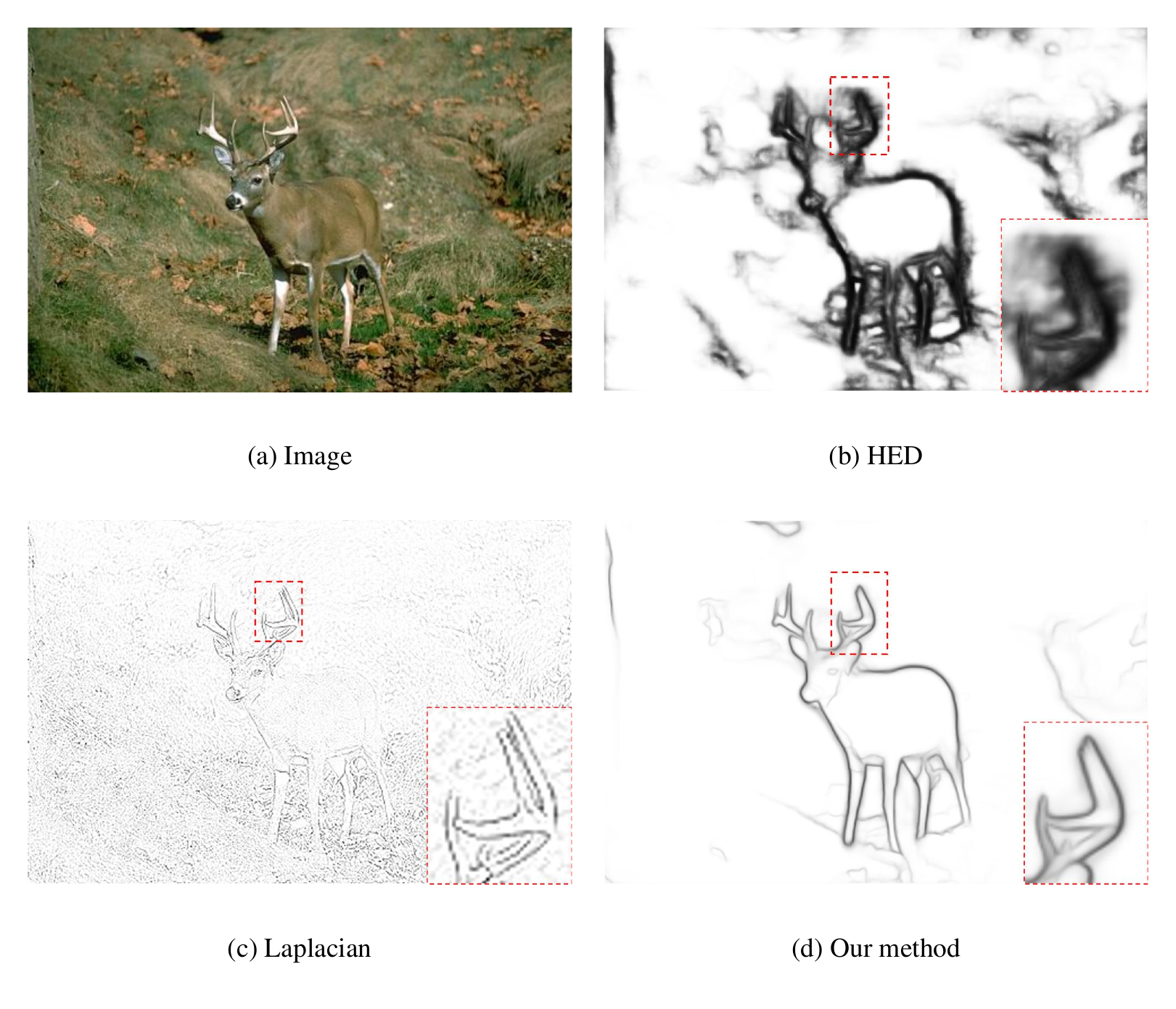}
    \caption{The crispness of edge maps from HED, Laplacian, and our method. (a) is an image from the BSDS500 dataset. (b) is the prediction of HED, which is thick and blurred. (c) is the output of Laplacian, which can leverage the second-order derivative information to generate crisp boundaries but with more noise. (d) is the output from our method, which is crisp and clean.}
    \label{example}
\end{figure}

However, there is still a major issue that needs to be addressed in edge detection: most modern DCNN-based models \cite{xie2015holistically,liu2017richer,he2019bi} are inclined to generate thick and noisy edge maps, such predictions introduce inaccuracies that compromise the spatial precision of high-level tasks, and it is frequently necessary to use a post-processing operation (NMS and morphological thinning) to obtain clean and precise edge maps. This issue can be decomposed into two subsidiary problems: (1) Excessive reliance on DCNN-based autonomous learning, without sufficient prior knowledge of edge information, which is essential for precise edge detection; (2) A highly imbalanced pixel distribution between edges and non-edges in an image, which presents difficulties in pixel classification.

In this work, we attempt to tackle this central issue by addressing the two subsidiary problems. Our goal is to make the network generate one-pixel width edge lines without using the post-processing operation. On the one hand, most DCNN-based methods \cite{xie2015holistically,liu2017richer,he2019bi,cao2020learning} leverage the powerful feature extraction capabilities of deep learning for edge detection. However, these methods neglect the utilization of prior knowledge regarding image edges, which are typically determined by abrupt changes in pixels. Image second-order derivative captures such changes and it precisely locates the positions of edge pixels by the zero-crossing points. Therefore, leveraging this characteristic can help DCNNs to locate the true edge pixels more accurately. In addition, the mere introduction of second-order derivative information can result in noisy edge maps, as illustrated in Fig. \ref{example} (c). In response to this circumstance, we combine the second-order derivative information with the multi-scale contextual information, developing a Second-order Derivative-based Multi-scale Contextual Enhancement Module (SDMCM). The SDMCM is based on the traditional Laplacian operator \cite{jain1995machine} which can provide image second-order derivative cues, and dilated convolution with a large receptive field to capture the long-range multi-scale contextual information which can filter out the noise. The proposed SDMCM empowers the network to attain higher precision in true edge pixel localization, consequently generating crisp and clean edge maps.

On the other hand, most DCNN-based approaches employ a weighted cross-entropy loss function \cite{xie2015holistically,liu2017richer,he2019bi,su2021pixel} to solve the issue of imbalanced pixel distribution, but such a strategy struggles to distinguish between true edge and false edge pixels, leading to the misclassification of false positive (FP) pixels as true positive (TP) pixels; this also causes the predictions with thick and noisy edges. Therefore, we propose a novel Hybrid Focal Loss (HFL) to control FPs and FNs, which can effectively reduce the impact of imbalanced pixel distribution. The HFL is based on the Tversky index \cite{salehi2017tversky} and focal loss \cite{lin2017focal} which constrains the network from image-level and pixel-level information, enhancing the accuracy in edge and non-edge pixel classification.

Furthermore, we employ the Conditional parameterized Convolution (CondConv) \cite{yang2019condconv} to construct a Boundary Refinement Module (BRM). CondConv can dynamically adjust convolutional kernel parameters according to the contextual information of input feature maps, and BRM exploits this characteristic to improve the model adaptability to different image scenarios, thereby further suppressing the interference and refining the predicted edge maps. In the end, we propose a LUS-Net which is based on the U-shape architecture for crisp edge detection, and the whole network can be split into three parts: encoder, skip-connections, and decoder. The encoder consists of a lightweight pre-trained model \cite{sandler2018mobilenetv2,ma2018shufflenet,tan2021efficientnetv2} to enable efficient prediction. The SDMCM serves as the skip-connection component. The decoder employs a dense connection to cascade each BRM, allowing the model to learn more diverse and rich feature representations. In addition, we conduct a series of experiments to demonstrate the effectiveness of our method, some instances are shown in Fig. \ref{example} (d).

In summary, the main contributions of our work can be summarized as follows:
\begin{itemize}
        \item [1.] We build a Second-order Derivative-based Multi-scale Contextual Enhancement Module (SDMCM), which can help our model locate the true edge pixels more accurately, consequently generating crisp and clean edge maps.
        \item [2.] We construct a Boundary Refinement Module (BRM) to further refine the edge maps, and propose a U-shape network named LUS-Net, which is based on the SDMCM and BRM, for crisp edge detection.
        \item [3.] We propose a novel Hybrid Focal Loss (HFL) based on Tversky index and focal loss, which can effectively alleviate the issue of imbalanced pixel distribution, resulting in suppressing misclassified false positive pixels near the true positive pixels.
	\item[4.] We conduct extensive experiments to demonstrate the advantages of our method and the results show that our method achieves the SOTA performance on three benchmark datasets.
\end{itemize}

The paper is structured as follows. Section \ref{Related work} presents a review of related work in edge detection. Section \ref{Methodology} provides an elaboration of the LUS-Net, which involves SDMCM, BRM, and HFL. Section \ref{Experiments} presents an analysis of our experiments. We evaluate the crispness of edges, provide detailed descriptions of the ablation study, analyze the function of each component, and compare our approach with recent state-of-the-art algorithms in edge detection. The final section, Section \ref{Discussion and conclusion}, summarizes our proposed method and discusses future directions.

\section{Related work}
\label{Related work}
Edge detection research spans over four decades, yielding extensive literature. In this section, we review some representative works which are split into two groups: traditional and deep learning-based methods.

\textbf{Traditional methods:} Early edge detection methods typically calculate image derivatives to produce edge maps.
The Sobel operator \cite{sobel1970camera} was another first-order edge detector that used two $3\times3$ kernels to calculate image gradients in horizontal and vertical directions. The Canny detector \cite{canny1986computational} was a robust algorithm that detected edges by reducing noise, calculating gradients, and applying non-maximum suppression and hysteresis thresholding. The Laplacian detector \cite{jain1995machine} located edge pixels by computing the second-order derivative of the image intensity, highlighting regions of rapid intensity change. Over time, some researchers improved edge detection by integrating texture, gradient, and other low-level features. Methods like gPb \cite{arbelaez2011contour} and SE \cite{dollar2014fast} used a classifier to generate object-level boundaries by utilizing these features. Despite their enhanced performance over derivative-based methods, they still relied on human-designed features and lacked semantic information, limiting further improvement.


\textbf{Deep learning-based methods:} Deep learning-based techniques significantly advanced the area of edge detection and played a crucial role in most state-of-the-art (SOTA) edge detection methods. SOTA edge detectors in recent years have mainly adopted convolutional neural networks. These methods achieved significant performance with higher F-scores and some of them even surpassed humans on benchmarks such as BSDS500 dataset. Xie et al. \cite{xie2015holistically} presented the first end-to-end edge detection architecture named Holistically-nested Edge Detection (HED), which was built on a fully convolutional VGG-16 network. They generated edge maps by fusing five side-output features with different scales and constructed a weighted cross-entropy loss function to address the problem of imbalanced pixel distribution. Based on HED, Liu et al. \cite{liu2017richer} further utilized multi-scale features by aggregating all features from different convolutional layers in each stage of VGG-16. This improved the network's ability to capture contextual information, making it the first to outperform humans on BSDS500. He et al. \cite{he2019bi} proposed an innovative bi-directional cascade network architecture, which leveraged information flow from shallow-to-deep and deep-to-shallow to capture multi-scale features within images comprehensively, significantly enhancing the performance. Soria et al. \cite{soria2023dense} proposed DexiNed, a network inspired by HED and Xception, which could produce detailed edge maps that were visually appealing without any pre-training or fine-tuning process. Su et al. \cite{su2021pixel} provided a lightweight yet effective solution for edge detection by integrating traditional edge detection operators into vanilla convolution in modern DCNN.

Regarding precise edge detection, various methods have contributed significantly to the field. Wang et al. \cite{wang2018deep} focused on guiding the network to predict sharp edge maps by employing sub-pixel convolution to upsample features. Deng et al. \cite{deng2018learning} explained the reason for edge thickness and proposed a new loss function based on the Dice coefficient which enables the network to generate crisp boundaries without requiring post-processing. 
Cao et al. \cite{cao2020learning} presented an innovative network by stacking refinement modules and implementing an adaptive weighting strategy for the optimal combination between different loss functions, resulting in a competitive performance. Huan et al. \cite{huan2021unmixing} presented a context-aware tracing strategy using a novel tracing loss and a context-aware fusion block for crisp edge detection. Xuan et al. \cite{xuan2022fcl} proposed a novel network architecture designed to enhance the accuracy of edge detection through fine-scale corrective learning mechanisms. Liu et al. \cite{liu2024generating} proposed GCB-Net, a U-shape encoder-decoder architecture with feature enhancement and fusion modules, which could generate crisp and clean edge maps through multi-scale features and a novel mixed loss function based on Tversky index. Pu et al. \cite{pu2022edter} EDTER, the first Transformer-based edge detector with a two-stage architecture. The model could extract well-defined boundaries of objects and meaningful edges by utilizing contextual information from the entire image, as well as detail local cues. Zhou et al. \cite{zhou2024muge} designed an edge granularity network to estimate edge granularity from annotations and integrated it into both spatial and frequency domains of feature maps, enabling the generation of diverse edge predictions ranging from coarse object contours to fine textures. Cetinkaya et al. \cite{cetinkaya2024ranked} constructed a ranking-based approach to address both class imbalance and label uncertainty in edge detection through ranking positive pixels over negatives and promoting high confidence edge pixels.

Despite their remarkable performance, existing methods still fall short in adequately addressing the challenge of edge thickness. Our approach is motivated by these pioneering works \cite{deng2018learning,salehi2017tversky,liu2024generating} to address the problem of edge thickness. Through the combination of multi-scale contextual and second-order derivative information, we enable the network to produce crisp and clean edge maps. Additionally, our method successfully addresses the issue of imbalanced pixel distribution by constructing a hybrid focal loss function, resulting in excellent performance.

\section{Methodology}
\label{Methodology}
\begin{figure}
    \centering
    \includegraphics[width=\textwidth]{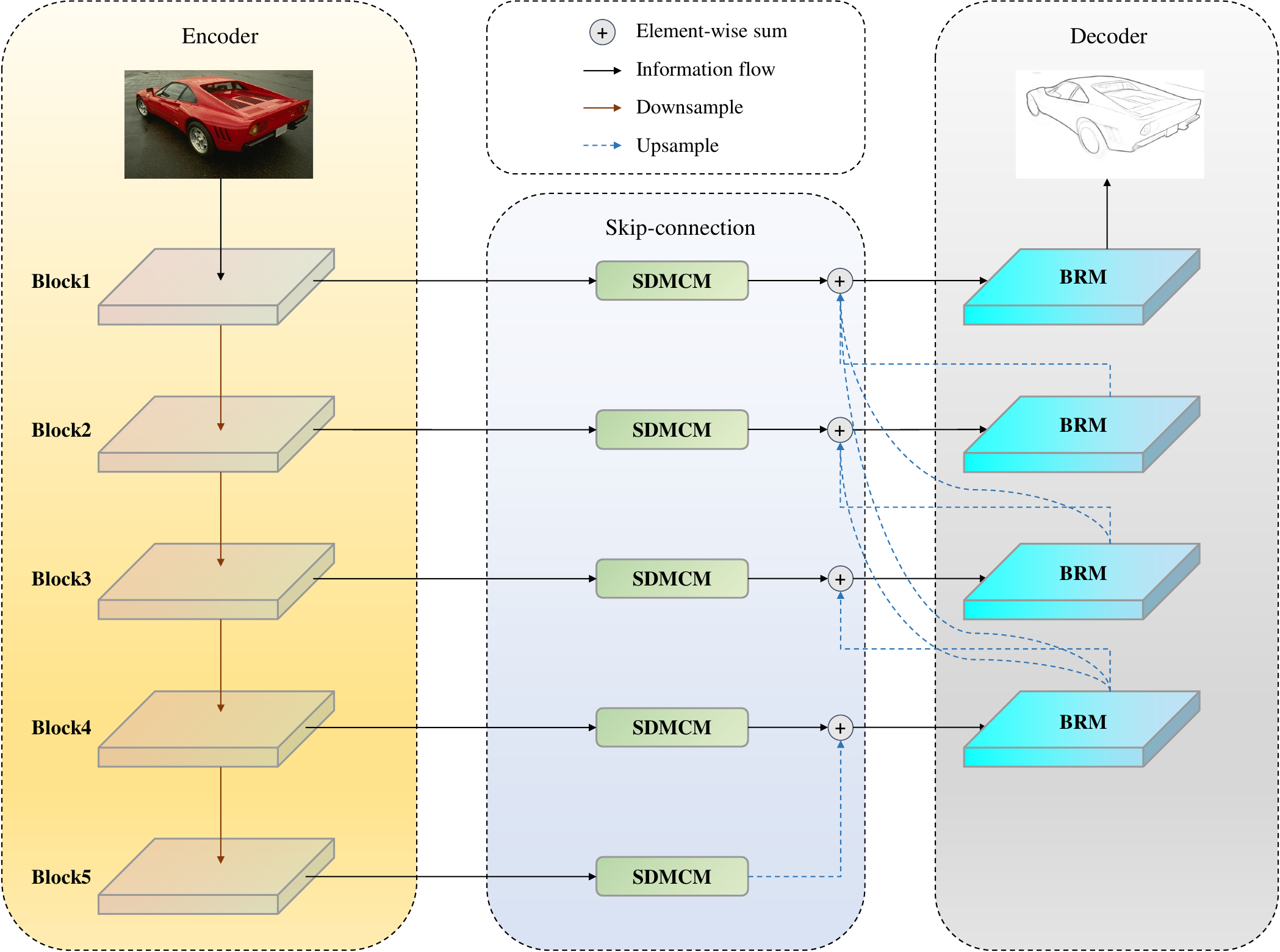}
    \caption{The architecture of LUS-Net.}
    \label{fig.1}
\end{figure}

In this section, we present the LUS-Net in detail. The whole network can be divided into three parts as shown in Fig. \ref{fig.1}: the top-down encoder, the skip-connection, and the bottom-up decoder. The top-down encoder comprises a lightweight pre-trained backbone, which can provide basic features with rich semantic information. The skip-connection consists of several Second-order Derivative-based Multi-scale Contextual Enhancement Modules (SDMCMs), and each SDMCM cascades to the corresponding stage of the backbone. The bottom-up decoder component is built on Boundary Refinement Modules (BRMs), which introduce Conditionally parameterized Convolution (CondConv) \cite{yang2019condconv} into edge detection for the first time. In particular, the decoder employs a dense connection to fuse different scale features, enhancing the expressive capability of the model. Additionally, our model is supervised using the Hybrid Focal Loss (HFL) function. We provide the details of each component in the following subsections.

\subsection{Lightweight pre-trained backbone}
\label{Lightweight pre-trained backbone}
Most SOTA edge detection methods adopt a pre-trained VGG \cite{simonyan2014very} or ResNet \cite{he2016deep} as a feature extraction backbone and then fine-tune them on the edge detection dataset. However, these backbones, with a large number of parameters, demand an expensive computational cost. To achieve efficient edge detection, we adopt and test three lightweight pre-trained backbones: MobileNetV2 \cite{sandler2018mobilenetv2}, ShuffleNetV2 \cite{ma2018shufflenet}, and EfficientNetV2 \cite{tan2021efficientnetv2}. These lightweight backbones have demonstrated comparable performance to VGG or ResNet networks while providing efficient inference capabilities. Specifically, these models employ a stride of two in the initial convolutional layer, resulting in excessive downsampling. This leads to feature maps with significantly reduced spatial resolution, damaging the quality of the generated contours. Therefore, we modify the stride from two to one. Additionally, we remove the classifier head to make the backbone network suitable for edge detection, which simultaneously reduces the network parameters.

\subsection{Second-order derivative-based multi-scale contextual enhancement module}
\label{SDMCM}
Image derivative information is used to locate edge pixels within an image, with commonly employed comprising first-order and second-order derivatives. The first-order derivative detects regions of rapid intensity change, indicating edges. The second-order derivative identifies the exact edge locations through zero-crossing points by measuring the rate of change of intensity. The information provided by the second-order derivative of an image is more precise than that of the first-order derivative because it is more sensitive to abrupt changes in pixel value, which typically corresponds to edge information. Given an input image $I$, a specific pixel location in the image is represented as $x_0$, the formula for computing the first-order derivative of the image in the x-direction at $x_0$ is:
\begin{equation}
    \frac{\partial I}{\partial x}\big|_{x=x_0}=I(x_0+1)-I(x_0)
\end{equation}
and the formula for the second-order derivative of the image at $x_0$ is:
\begin{equation}
    \frac{\partial^2f}{\partial x^2}\big|_{x=x_0}=I(x_0+1)+I(x_0-1)-2I(x_0)
\end{equation}

\begin{figure}[htbp]
    \centering
    \includegraphics[width=\textwidth]{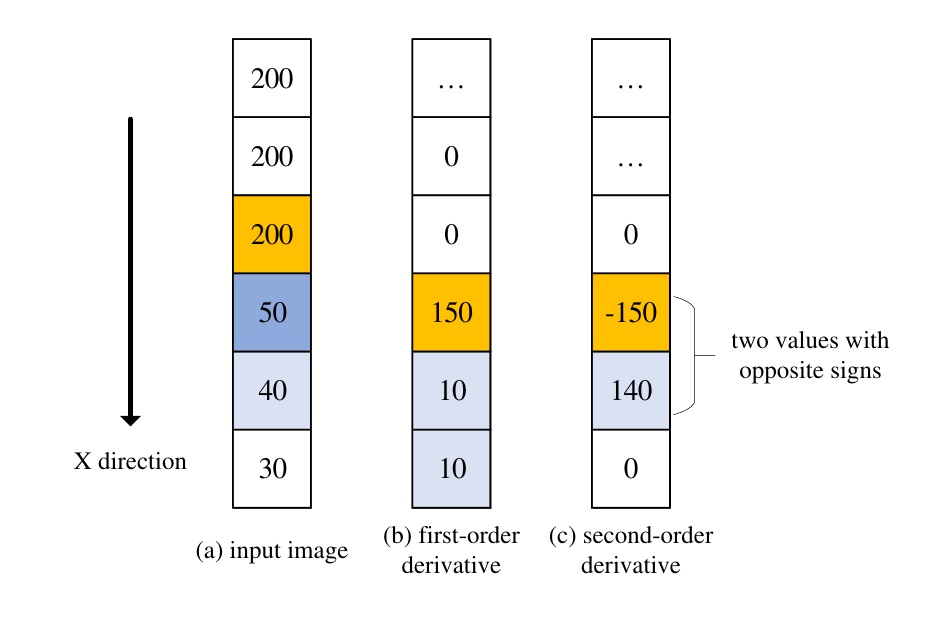}
    \caption{The process of image first-order derivative and second-order derivative.}
    \label{Derivative}
\end{figure}

The process of these two derivatives is shown in Fig. \ref{Derivative}. For the first-order derivative, pixel values remain constant and have the same sign in areas of slow variation, changing only at sharp grayscale transitions. In contrast, the second-order derivative is zero in slow variation regions but shows opposite signs at abrupt changes, creating a one-pixel width edge, which is precisely the desired output we require. This accurately locates edge pixels, as the opposite signs enhance edge contrast, making the second-order derivative more effective for precise edge detection.

\begin{figure}[htbp]
    \centering
    \includegraphics[scale=0.6]{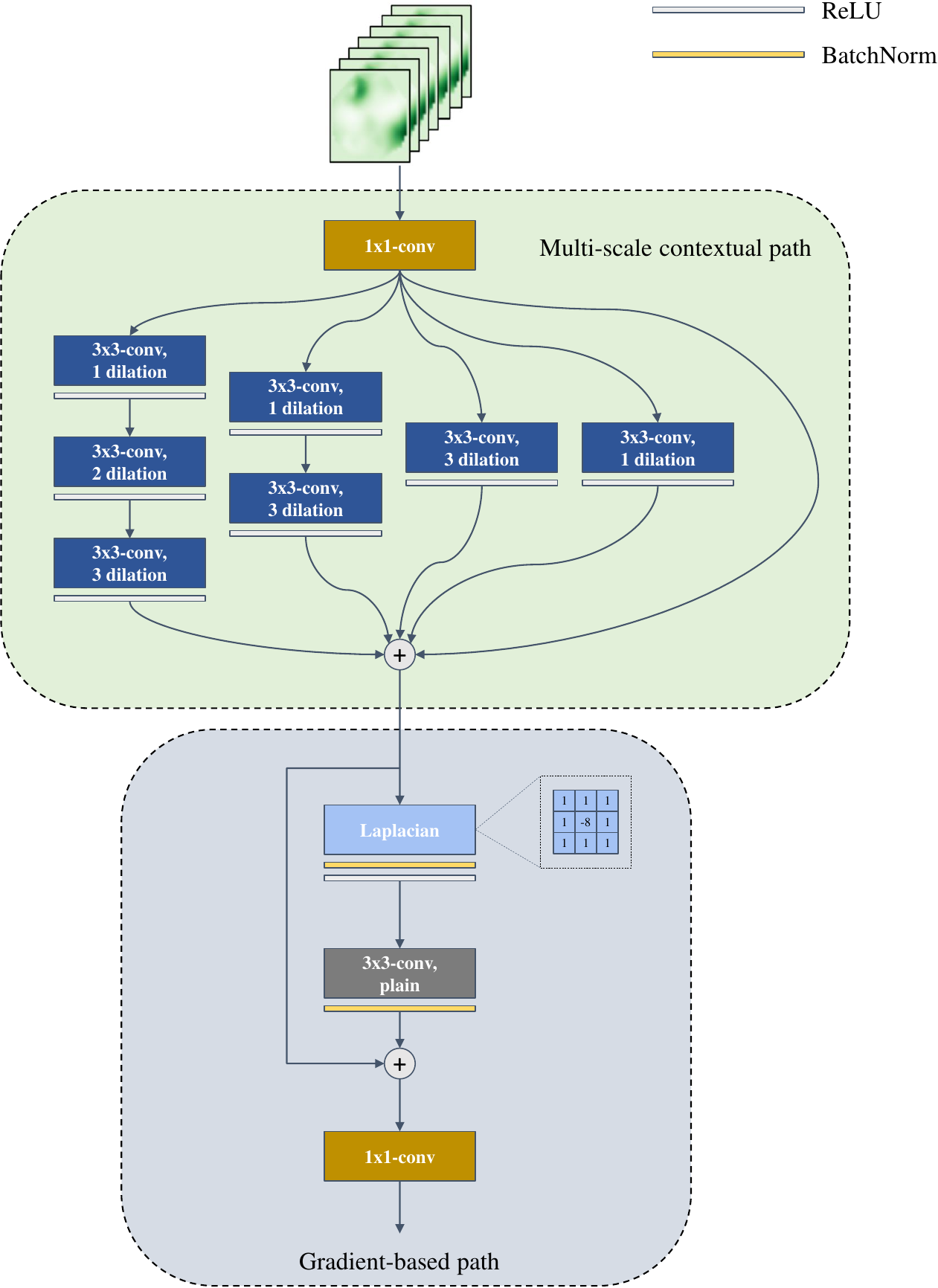}
    \caption{The diagram of SDMCM.}
    \label{fig.2}
\end{figure}

Relying solely on the second-order derivative for edge detection is insufficient because of its sensitivity to noise, which amplifies noise influence despite accurately locating edge pixels. To mitigate this, we propose a second-order derivative-based multi-scale contextual enhancement module (SDMCM). By introducing multi-scale contextual information, the receptive field expands, providing long-range semantic and structural information that helps the network distinguish true edge pixels from noise. True edge pixels are associated with objects or structures, while noise edges lack semantic coherence. Leveraging this information, the network can suppress noise edges and generate crisp edge maps, effectively addressing the thickness issue. The architecture of SDMCM can be seen in Fig. \ref{fig.2}, and it can be divided into two parts: the multi-scale contextual path and the second-order derivative path.

In the multi-scale contextual path, we first employ a channel compression operation $\mathcal{C}(\cdot)$ which is a $1\times1$ convolution to compress the input feature channel. This operation effectively prevents overfitting while simultaneously reducing the number of parameters. The compression ratio $r\in\left\{\frac{1}{2},\frac{1}{4},\frac{1}{8}\right\}$, we find that changes in compression rate have an impact on the crispness of the predicted edge maps (Section \ref{Experiments} provides details). After the channel compression, we construct four parallel branches $\left\{\mathcal{B}_1(\cdot), \mathcal{B}_2(\cdot), \mathcal{B}_3(\cdot), \mathcal{B}_4(\cdot)\right\}$ to produce multi-scale contextual features, and each branch consists of several $3\times3$ dilated Conv-ReLU sequences. Introducing dilated convolution using a fixed dilated rate can broaden the receptive field, however, it also produces fixed feature intervals, which leads to blurred and discontinuous edges. Therefore, we stack a series of dilated convolutions with different dilated rates $d\in\left\{1,2,3\right\}$, allowing for filling intervals in the features. In this way, we can utilize all the features in continuous convolution kernels, without producing unused features or fixed intervals, and capturing long-range semantic information by expanding the receptive field. In the end, we combine the features from four branches by element-wise sum to obtain the mixed features which contain rich multi-scale contextual information. 

We employ the original Laplacian template to build the second-order derivative path, which consists of a Laplacian-Batchnorm-ReLU sequence, a $3\times3$ Conv-Batchnorm sequence, and a $1\times1$ convolution. Additionally, we introduce a shortcut connection in both two paths to more effectively model complex non-linear transformations, thereby enhancing the expressive power of the model. In this situation, for an input two-dimensional feature map $X\in\mathcal{R}^{H\times W}$, the output feature map $Y\in\mathcal{R}^{H\times W}$ is obtained by
\begin{equation}
\begin{aligned}
    Y=&\mathcal{S}\left[\mathcal{B}_1\left(\mathcal{C}(X)\right)+\mathcal{B}_2\left(\mathcal{C}(X)\right)+\mathcal{B}_3\left(\mathcal{C}(X)\right)+\mathcal{B}_4\left(\mathcal{C}(X)\right)+\mathcal{C}(X)\right]
\end{aligned}
\end{equation}
our newly developed SDMCM significantly enhances the capability of the model to locate the true edge pixels by combining the multi-scale contextual information and the second-order derivative information, consequently making our model generate crisp and precise edge maps.

\subsection{Boundary refinement module}
\label{BRM}
\begin{figure}[htbp]
    \centering
    \includegraphics[scale=0.8]{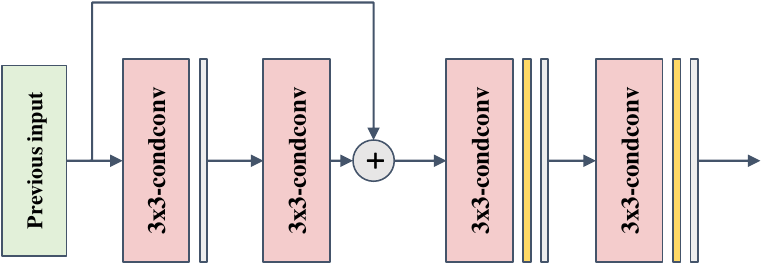}
    \caption{The structure of BRM.}
    \label{fig.3}
\end{figure}

We develop a novel boundary refinement module (BRM) as shown in Fig. \ref{fig.3}, which is based on the Conditionally parameterized Convolution \cite{yang2019condconv} (CondConv) to further refine the boundaries. CondConv can improve the model adaptability without significantly increasing the number of parameters by consolidating multiple convolutions into a single conditionally parameterized convolutional kernel. We leverage the unique characteristics of CondConv to build the BRM. This module can address the issue of illumination variations in an image, which leads to generating spurious edges.

The BRM consists of a residual block and two $3\times3$ CondConv-Batchnorm-ReLU sequences. The residual block effectively facilitates information flow by summing operation, and two $3\times3$ CondConv-Batchnorm-ReLU sequences further refine the features from the residual block, thereby improving the performance in edge detection. Our BRM can learn the contextual features of image illumination disturbances and modulate the convolutional kernel parameters accordingly, which aids in suppressing the spurious edge responses in these regions, resulting in more refined and precise edge maps. By harnessing the adaptive nature of CondConv, the BRM allows the model to adjust its convolutional behavior according to the input context dynamically, mitigating the impact of interference factors, and generating accurate and semantically consistent edge representations.

Additionally, we utilize $3\times3$ depthwise separable convolution coupled with bilinear interpolation to upsample the resolution of feature maps from BRM. This combination enables us to increase the spatial dimensions of the feature maps while preserving important details and spatial relationships. In the end, we employ a dense connection to cascade each BRM, thereby integrating different scale features and allowing the model to fully leverage high-level semantic information to filter out the noise pixels.

\subsection{Hybrid focal loss}
\label{loss}
As a basic computer vision task, edge detection encounters a significant challenge of imbalanced class distribution, with edge pixels typically comprising only about 10\% of an image. While current mainstream methods use weighted cross-entropy loss to tackle this issue, they often result in thick boundaries as such strategy lacks global structural information \cite{deng2018learning}. Therefore, edge detection demands more complete solutions beyond simple class balancing, as it requires both pixel-level precision and global structural information. Inspired by the focal loss \cite{lin2017focal} and Tversky index \cite{salehi2017tversky}, we propose a hybrid approach that specifically addresses these unique challenges. Our key insight is that effective edge detection requires simultaneous optimization of both local pixel-level accuracy and global structural information, which neither existing loss function alone can achieve.

The focal loss aims to tackle the issue of imbalanced data sample distribution, and we adopt this concept to tackle the imbalanced pixel distribution in an image. While standard cross-entropy has been widely used in pixel-level binary classification to measure prediction accuracy, it exhibits significant limitations when dealing with imbalanced distributions. Specifically, the loss function becomes predominantly influenced by easily classified majority-class pixels, leading to insufficient learning of the more challenging minority-class pixels. Therefore, a focusing parameter $\gamma \geq 0$ is introduced into cross-entropy to downweigh the loss for well-classified pixels and focus training on misclassified pixels. The focal loss can be written as:
\begin{equation}
\begin{aligned}
    L_{FL}=-\alpha\sum_{i=1}^{N}\left(\left(1-p_{i}\right)^\gamma g_{i} \log p_{i}+p_{i}^\gamma\left(1-g_{i}\right)\log\left(1-p_{i}\right)\right)
\end{aligned}
\end{equation}
where $p_{i}$ and $g_{i}$ represent the value of $i$-th pixel on a predicted edge map and its corresponding groundtruth image, respectively. $N$ is the total number of pixels in an image. $(1-p_{i})^\gamma$ is a modulating factor and $\alpha$ is a balance factor for positive and negative pixels. When $\gamma \geq 0$, the modulating factor $(1-p_{i})^\gamma$ produces an adaptive scaling effect. When $p_i$ approaches 1, the factor reduces the loss for well-classified pixels. Conversely, when $p_i$ is small, the factor maintains high loss for misclassified pixels. This adaptive modulation mechanism ensures that the model pays more attention to the minority class and improves its ability to correctly classify these challenging cases, providing a principled solution to the imbalanced pixel distribution.

Although focal loss can address class imbalance and achieve correct pixel classification, it primarily operates on local pixel relationships, neglecting global structural information. This limitation impairs its ability to suppress false edge pixels around true edge pixels, often resulting in thick edge maps. To address this issue, we introduce global structural information through the Tversky index, which is formally defined as follows:
\begin{equation}
\begin{aligned}
    T\left(P,G;\alpha,\beta\right)=\frac{|PG|}{|PG|+\alpha|P/G|+\beta|G/P|}
\end{aligned}
\end{equation}
where P indicates the predicted edge maps and G indicates its corresponding labels. $\alpha$ and $\beta$ control the weight of false negatives (FNs) and false positives (FPs), respectively. The sum of these two parameters is set to 1.

Inspired by the focal loss modification of the cross-entropy loss, we adopt a similar strategy and introduce $\gamma$ into the reciprocal of Tversky index. The focal Tversky loss function can be defined as:
\begin{equation}
\begin{aligned}
    L_{FT}=\left(\frac{\sum_{i=1}^{N} p_{i} g_{i}+\alpha \sum_{i=1}^{N} \left({{p_{i} (1-g_{i})}}\right)^2 +\beta \sum_{i=1}^{N} \left({(1-p_{i}) g_{i}}\right)^2 +C} {\sum_{i=1}^{N} p_{i} g_{i} +C}\right)^\gamma
\end{aligned}
\end{equation}
where $p_i(1-g_i)$ and $(1-p_i)g_i$ represent FPs and FNs, respectively. $C=1\times 10^{-7}$ is a constant number to prevent the numerator/denominator from being 0. 

\begin{figure}[htbp]
    \centering
    \includegraphics[scale=0.35]{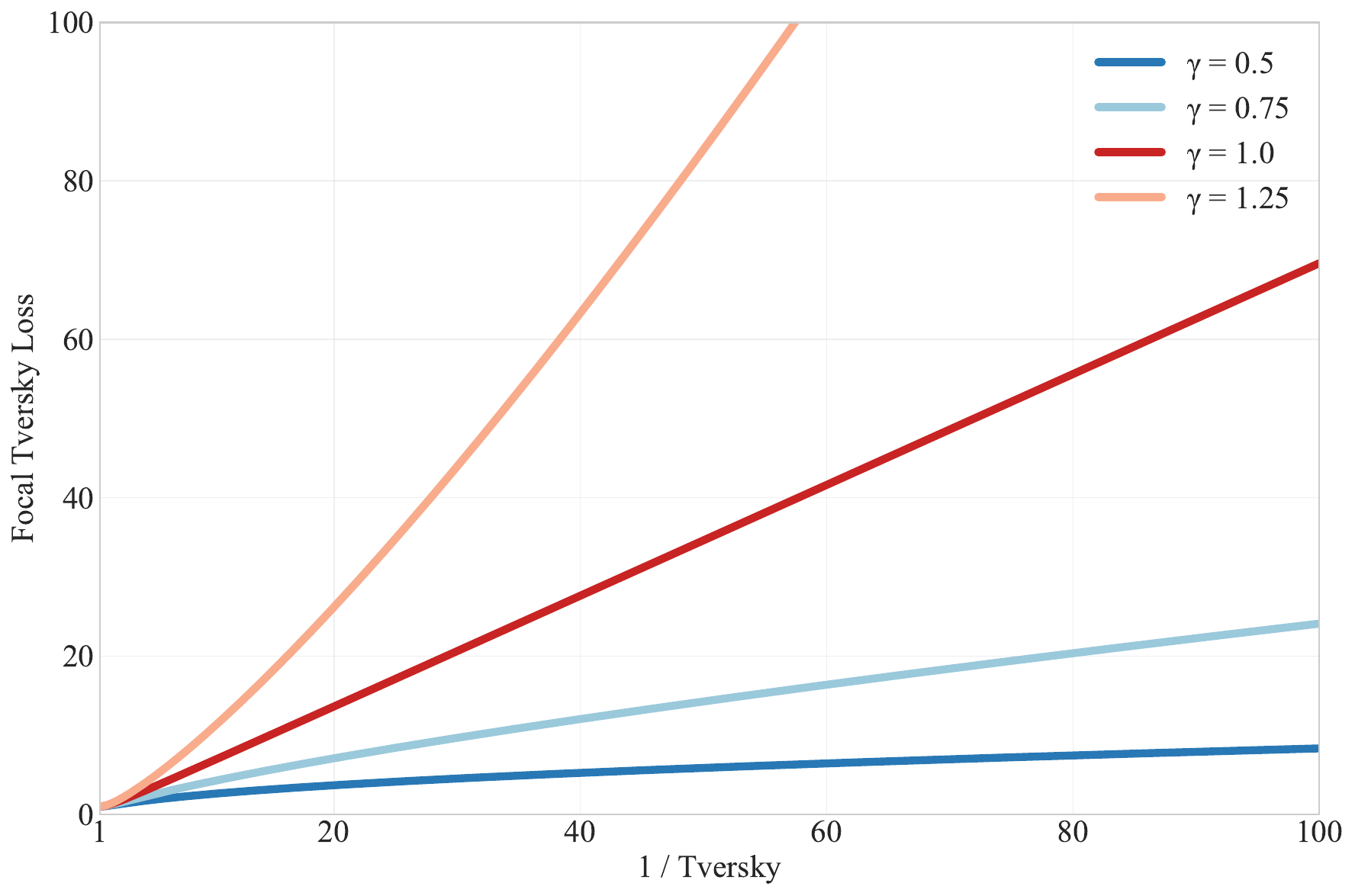}
    \caption{Visualization of focal Tversky loss curves under different $\gamma$.}
    \label{Tversky_gamma}
\end{figure}

Here, $\gamma=0.75$ serves as a focusing parameter. Specifically, when the $\frac{1}{Tversky}\gg$ 1, suggesting predictions with errors, $\gamma$ reduces the loss significantly. Conversely, when the $\frac{1}{Tversky}$ is close to 1, indicating accurate pixel classification, the loss shows high fluctuation. Therefore, during early training, as the network lacks global structural information, it emphasizes pixel-level predictions. The reduction effect of $\gamma$ enables the focal loss to dominate the primary learning process. As training process to later stages, most pixels are well-classified, and the $\frac{1}{Tversky}$ approaches 1. As evidenced by Fig. \ref{Tversky_gamma}, the loss function exhibits steeper gradients near $\frac{1}{Tversky}$ close to 1, focal Tversky loss dominates the learning process at this point, enabling the model to focus more attention on these well-classified pixels for further refinement by utilizing the global structural information.

Our hybrid focal loss function is a weighted fusion of the above two loss functions, which can be defined as:
\begin{equation}
\begin{aligned}
    L_{HFL}=L_{FT}+\lambda L_{FL}
\end{aligned}
\end{equation}
where we set $\lambda=0.001$ to balance the weight between focal loss and focal Tversky loss. The focal loss operates at the pixel level, using adaptive scaling effect to address class imbalance, while the focal Tversky loss operates at the image level, leveraging global structural information for accurate pixel location. Therefore, our hybrid focal loss function can tackle both class imbalance issues and thickness problems, consequently improving the performance of the model.

\section{Experiments}
\label{Experiments}
In this section, we provide a comprehensive account of the implementation details, encompassing hyperparameters, the adopted datasets, and their augmentation strategy. We then introduce the two evaluation methods employed in this work, followed by a series of ablation experiments on our approach. Finally, we compare our proposed method with some SOTA edge detection algorithms and demonstrate its superiority.

\subsection{Implementation details}
\label{Implementation details}
Our network is built using the Pytorch deep learning framework. During the training phase, the hyperparameters are as follows: the mini-batch size is 8, the initial learning rate is $1\times 10^{-4}$, the learning rate decay is 0.1, the weight decay is $5\times 10^{-4}$ and the number of training epochs is 40. We decay the learning rate every 5 epochs and adopt the Adam method for optimization. All the experiments are performed using a single Tesla A40 GPU.

In addition, we test our method on three benchmarks: BSDS500 \cite{arbelaez2011contour}, NYUD-V2 \cite{silberman2012indoor}, and BIPED \cite{soria2023dense}. The Berkeley Segmentation Dataset and Benchmarks 500 (BSDS500) is a comprehensive dataset that has been extensively utilized in the field of computer vision, particularly for evaluating edge detection algorithms. This dataset encompasses a diverse collection of 500 natural images, which are further divided into a standard split of 200 training, 100 validation, and 200 testing images. Each image in the BSDS500 is meticulously annotated by multiple human annotators (usually around 5 to 7), providing a rich set of groundtruth images that represents a wide range of human perception in terms of object boundaries. The NYU Depth Dataset V2 (NYUD-V2) is an influential benchmark that includes depth information and primarily consists of indoor scene images. This dataset comprises 1449 pairs of images, with each pair consisting of RGB and depth images. It is divided into three subsets: a training subset containing 381 images, a validation subset containing 414 images, and a testing subset comprising 654 images. The depth information can be encoded into three channels: horizontal disparity, height above ground, and angle with gravity (HHA). This allows for the storage of depth information in a three-channel RGB image, known as the HHA feature image. The Barcelona Images for Perceptual Edge Detection (BIPED) dataset is a high-quality dataset for evaluating perceptual edge detection algorithms. It contains 250 high-resolution images ($1280\times720$ pixels) captured in outdoor scenes. The images are divided into a training set of 200 images and a test set of 50 images. During the training process, we merge the training and validation subsets from BSDS500 and NYUD-V2 into a single set, respectively. As for BIPED, we adopt their data settings.

As for data augmentation, we follow the previous works \cite{xie2015holistically,liu2017richer,deng2018learning}, crop and flip the image-label pairs by randomly rotating 24 angles, and all the three datasets employ the same augment strategy.

\subsection{Evaluation methods}
\label{Evaluation methods}
To evaluate the quality of generated edge maps, we report the F-score $\left(\frac{2\times Precision\times Recall}{Precision+Recall}\right)$ and Average Precision (AP), which are widely used in edge detection \cite{xie2015holistically,liu2017richer,deng2018learning}. Specifically, $Precision=\frac{TP}{TP+FP}$ and $Recall=\frac{TP}{TP+FN}$, where $TP$, $FP$, and $FN$ represent the number of correctly classified edge pixels, the number of incorrectly classified edge pixels, and the number of missed edge pixels, respectively. Once the network has generated edge maps, the primary step is to apply a threshold to convert them into binary edge maps, and then match these to the groundtruth images to calculate the F-score. There are two choices to set this threshold: optimal dataset scale (ODS) and optimal image scale (OIS). ODS F-score is calculated across the whole dataset by selecting a fixed threshold that maximizes the F-score for all images collectively. OIS F-score is calculated for each image independently, choosing the best threshold for that particular image to maximize its F-score. AP summarizes the Precision-Recall curve by calculating the area under this curve, thereby providing a single score that captures the model's performance across all possible thresholds. In addition, we also report the number of parameters (Params) and Frames Per Second (FPS) to evaluate the efficiency of the model.

To comprehensively evaluate the crispness of edge maps, following these previous works \cite{wang2018deep,deng2018learning}, we adopt the same strategy to evaluate our method including standard evaluation (S-Eval) and crispness evaluation (C-Eval).

\textbf{Standard evaluation method.} This method first applies standard post-processing process (NMS and morphological thinning) to the predicted edge maps, then matches these post-processed results with groundtruth images to calculate ODS, OIS, and AP. The S-Eval aims to assess the correctness of predicted edge maps.

\textbf{Crispness evaluation method.} To evaluate whether the network can directly generate clean and crisp edge maps, we calculate ODS, OIS, and AP on the raw predictions without such standard post-processing steps.

\subsection{Ablation experiments on BSDS500 dataset}
\label{Ablation experiment}
To analyze the effectiveness of each configuration in our LUS-Net, we conduct a series of ablation experiments on the BSDS500 dataset with MobileNetV2 as the encoder. The performance is evaluated using both S-Eval and C-Eval methods.

\textbf{Loss function:} Firstly, we perform ablation experiments on various loss functions to demonstrate the effectiveness of our HFL. To identify the optimal configuration of HFL, we conduct a series of comparison experiments with different parameters, and the comparison results are shown in rows 1-6 of Table \ref{Tab1}. It can be observed that, as $\lambda$ decreases from 1 to 0.001, there is a notable increase in C-Eval metrics, while S-Eval metrics maintain relatively stable performance. Such observation can be explained by the complementary roles of the two loss components: focal Tversky loss serves to regulate the crispness of edge delineation, whereas focal loss ensures the accuracy of pixel-level classification in the predicted edge maps. Therefore, the optimal value of $\lambda$ is 0.001. Additionally, when fixed $\lambda=0.001$, the best results of both S-Eval and C-Eval are obtained at $\beta=0.7$, outperforming other values (0.6 and 0.8). In rows 7-10, it can be observed that the best performance is obtained at $\gamma=0.75$.
Furthermore, we compare our HFL loss with other loss functions, including Weighted Cross-Entropy \cite{xie2015holistically} (WCE), Hybrid Quality Focal Loss \cite{li2022generalized} (HQFL), and Hybrid Varifocal Loss \cite{zhang2021varifocalnet} (HVFL). The comparison results shown in rows 11-14 demonstrate that the HFL maintains competitive performance in S-Eval (ODS=0.805, OIS=0.827, AP=0.805). More importantly, HFL demonstrates superior performance across all metrics in C-Eval (ODS=0.698, OIS=0.705, AP=0.754), notably outperforming WCE, HQFL, and HVFL. These results strongly indicate that our proposed HFL effectively enhances the model's capability to produce crisp and clean edge maps while maintaining robust performance in pixel-level classification.

\begin{table}[htbp]
	\caption{Ablation experiment results about different configurations in our network.}
	\label{Tab1}
	\centering
	\resizebox{\textwidth}{!}{
        \begin{tabular}{c|c|c|c|ccc|ccc}
		\toprule
		\multirow{2}{*}{Method}& \multirow{2}{*}{Compression ratio}& \multirow{2}{*}{Loss}& \multirow{2}{*}{Derivative}& \multicolumn{3}{c|}{S-Eval}& \multicolumn{3}{c}{C-Eval}\\
        \cline{5-10}
        \multirow{20}{*}{LUS-Net-M}& & & &ODS& OIS& AP& ODS& OIS& AP\\
        \hline
        & 0.25& HFL$\left(\beta=0.7,\lambda=1\right)$& Laplacian& 0.805& 0.825& 0.807& 0.686& 0.694& 0.742\\
        & 0.25& HFL$\left(\beta=0.7,\lambda=0.1\right)$& Laplacian& 0.807& 0.827& 0.807& 0.685& 0.694& 0.744\\
        & 0.25& HFL$\left(\beta=0.7,\lambda=0.01\right)$& Laplacian& 0.809& 0.828& 0.808& 0.691& 0.696& 0.750\\
        & \textbf{0.25}& \textbf{HFL$\left(\beta=0.7,\lambda=0.001\right)$}& \textbf{Laplacian}& \textbf{0.805}& \textbf{0.827}& \textbf{0.805}& \textbf{0.698}& \textbf{0.705}& \textbf{0.754}\\
        & 0.25& HFL$\left(\beta=0.6,\lambda=0.001\right)$& Laplacian& 0.802& 0.822& 0.803& 0.685& 0.691& 0.740\\
        & 0.25& HFL$\left(\beta=0.8,\lambda=0.001\right)$& Laplacian& 0.803& 0.824& 0.804& 0.689& 0.694& 0.740\\
        \cline{2-10}
        & 0.25& HFL$\left(\gamma=0.5,\lambda=0.001\right)$& Laplacian& 0.807& 0.825& 0.814& 0.682& 0.690& 0.742\\
        & \textbf{0.25}& \textbf{HFL$\left(\gamma=0.75,\lambda=0.001\right)$}& \textbf{Laplacian}& \textbf{0.805}& \textbf{0.827}& \textbf{0.805}& \textbf{0.698}& \textbf{0.705}& \textbf{0.754}\\
        & 0.25& HFL$\left(\gamma=1.0,\lambda=0.001\right)$& Laplacian& 0.806& 0.826& 0.814& 0.675& 0.691& 0.728\\
        & 0.25& HFL$\left(\gamma=1.25,\lambda=0.001\right)$& Laplacian& 0.805& 0.827& 0.821& 0.682& 0.689& 0.735\\
        \cline{2-10}
        & 0.25& WCE& Laplacian& 0.812& 0.833& 0.850& 0.693& 0.702& 0.745\\
        & \textbf{0.25}& \textbf{HFL$\left(\beta=0.7,\lambda=0.001\right)$}& \textbf{Laplacian}& \textbf{0.805}& \textbf{0.827}& \textbf{0.805}& \textbf{0.698}& \textbf{0.705}& \textbf{0.754}\\
        & 0.25& HQFL$\left(\beta=0.7,\lambda=0.001\right)$& Laplacian& 0.803& 0.821& 0.808& 0.682& 0.694& 0.727\\
        & 0.25& HVFL$\left(\beta=0.7,\lambda=0.001\right)$& Laplacian& 0.805& 0.825& 0.804& 0.690& 0.698& 0.753\\
        \cline{2-10}
        & 0.125& HFL$\left(\beta=0.7,\lambda=0.001\right)$& Laplacian& 0.801& 0.821& 0.810& 0.682& 0.687& 0.739\\
        & \textbf{0.25}& \textbf{HFL$\left(\beta=0.7,\lambda=0.001\right)$}& \textbf{Laplacian}& \textbf{0.805}& \textbf{0.827}& \textbf{0.805}& \textbf{0.698}& \textbf{0.705}& \textbf{0.754}\\
        & 0.5& HFL$\left(\beta=0.7,\lambda=0.001\right)$& Laplacian& 0.802& 0.824& 0.798& 0.692& 0.698& 0.741\\   
        \cline{2-10}
        & \textbf{0.25}& \textbf{HFL$\left(\beta=0.7,\lambda=0.001\right)$}& \textbf{Laplacian}& \textbf{0.805}& \textbf{0.827}& \textbf{0.805}& \textbf{0.698}& \textbf{0.705}& \textbf{0.754}\\
        & 0.25& HFL$\left(\beta=0.7,\lambda=0.001\right)$& Sobel& 0.803& 0.824& 0.805& 0.690& 0.697& 0.745\\
        & 0.25& HFL$\left(\beta=0.7,\lambda=0.001\right)$& Scharr& 0.803& 0.822& 0.803& 0.687& 0.693& 0.742\\
	\bottomrule
	\end{tabular}
    }
\end{table}

\begin{figure}[htbp]
    \centering
    \includegraphics[scale=0.55]{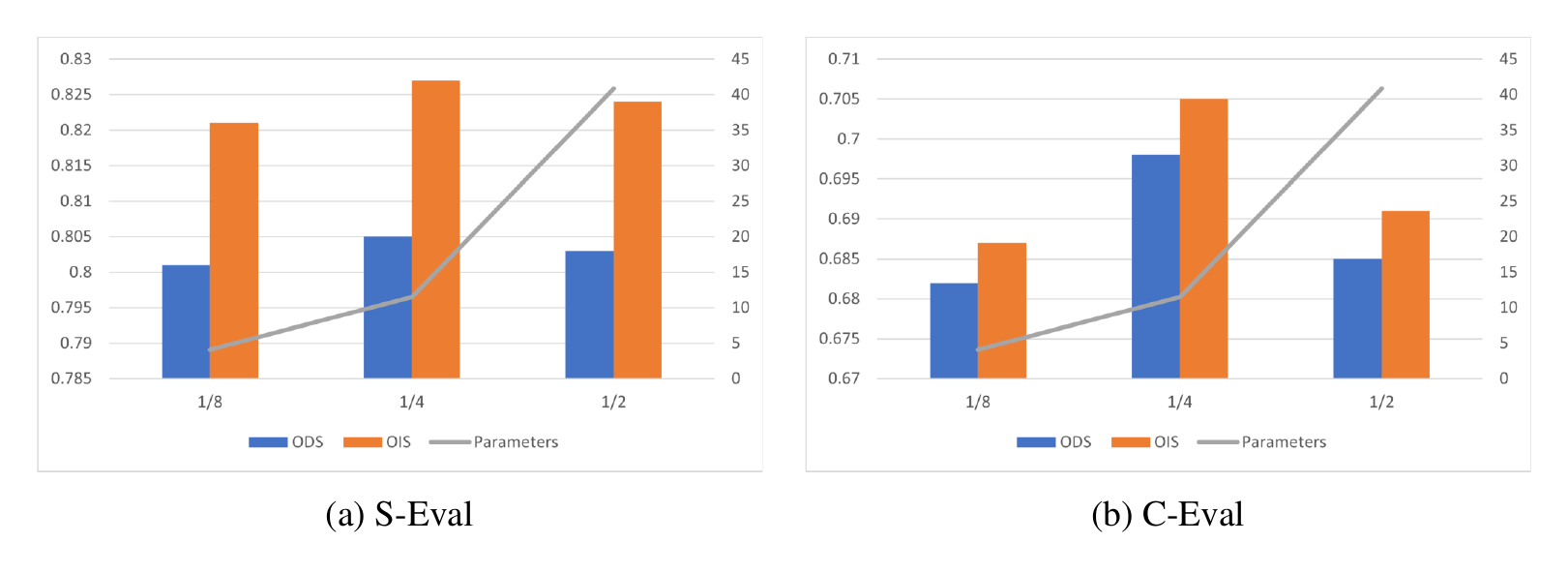}
    \caption{The effect of different compression ratios on network performance.}
    \label{ratio}
\end{figure}

\textbf{Compression ratio:} We also explore the effect of different channel compression ratios $r$ in SDMCM on the model performance and the experiment results are shown in rows 15-17 of Table \ref{Tab1} and Fig. \ref{ratio}. As the ratio increases, the number of model parameters also increases, resulting in a higher computational cost. However, the performance does not improve with an increasing compression ratio. The best performance is obtained at $r=\frac{1}{4}$. The reason for this phenomenon is that a higher compression ratio causes overfitting, resulting in a lower performance both in S-Eval and C-Eval. Therefore, we set the compression ratio to $\frac{1}{4}$, which achieves a balance between performance and parameters.

\textbf{Image derivative information:} In addition, we investigate the effectiveness of different derivative operators by replacing the Laplacian operator in SDMCM with first-order derivative operators (Sobel and Scharr). As shown in rows 18-20 of Table \ref{Tab1}, both alternative operators lead to consistent performance degradation across all metrics, substantiating our theoretical analysis that second-order derivative information provides more discriminative features for edge detection compared to first-order derivatives.

\textbf{Network components:} Finally, we conduct the last ablation studies to evaluate the contribution of each proposed component within our network, and the results are presented in Table \ref{Tab2}. The experiment results demonstrate that our complete model with all components achieves the best performance. When removing each component from our model, the performance shows varying degrees of degradation in both S-Eval and C-Eval. Notably, the exclusion of BRM leads to the most significant performance drop, underscoring its crucial role in our framework. These comprehensive experiments convincingly demonstrate the effectiveness and necessity of each proposed component.

\begin{table}[htbp]
	\caption{Ablation experiment results of each network component. DC indicates the dense connection.}
	\label{Tab2}
	\centering
	\begin{tabular}{@{}c|ccc|ccc|ccc@{}}
		\toprule
		\multirow{2}{*}{Methods}& \multirow{2}{*}{SDMCM}& \multirow{2}{*}{BRM}& \multirow{2}{*}{DC}& \multicolumn{3}{c|}{S-Eval}& \multicolumn{3}{c}{C-Eval}\\
        \cline{5-10}
        &&&&ODS& OIS& AP& ODS& OIS& AP\\
		\hline
		\multirow{4}{*}{LUS-Net-M}& \ding{51}& \ding{51}& \ding{51}& \textbf{0.805}& \textbf{0.827}& \textbf{0.805}& \textbf{0.698}& \textbf{0.705}& \textbf{0.754}\\
		& \ding{55}& \ding{51}& \ding{51}& 0.803& 0.822& 0.801& 0.686& 0.692& 0.736\\
        & \ding{51}& \ding{55}& \ding{51}& 0.794& 0.813& 0.791& 0.647& 0.654& 0.675\\
        & \ding{51}& \ding{51}& \ding{55}& 0.804& 0.825& 0.811& 0.687& 0.693& 0.739\\
		\bottomrule
	\end{tabular}
\end{table}

\subsection{Comparison with SOTA methods}
\label{Comparison with SOTA methods}
In this subsection, we compare our algorithm with current SOTA methods by conducting an experiment using three datasets: BSDS500 \cite{arbelaez2011contour}, NYUD-V2 \cite{silberman2012indoor}, and BIPED \cite{soria2023dense}.

\begin{figure}[htbp]
	\centering
	\includegraphics[width=\textwidth]{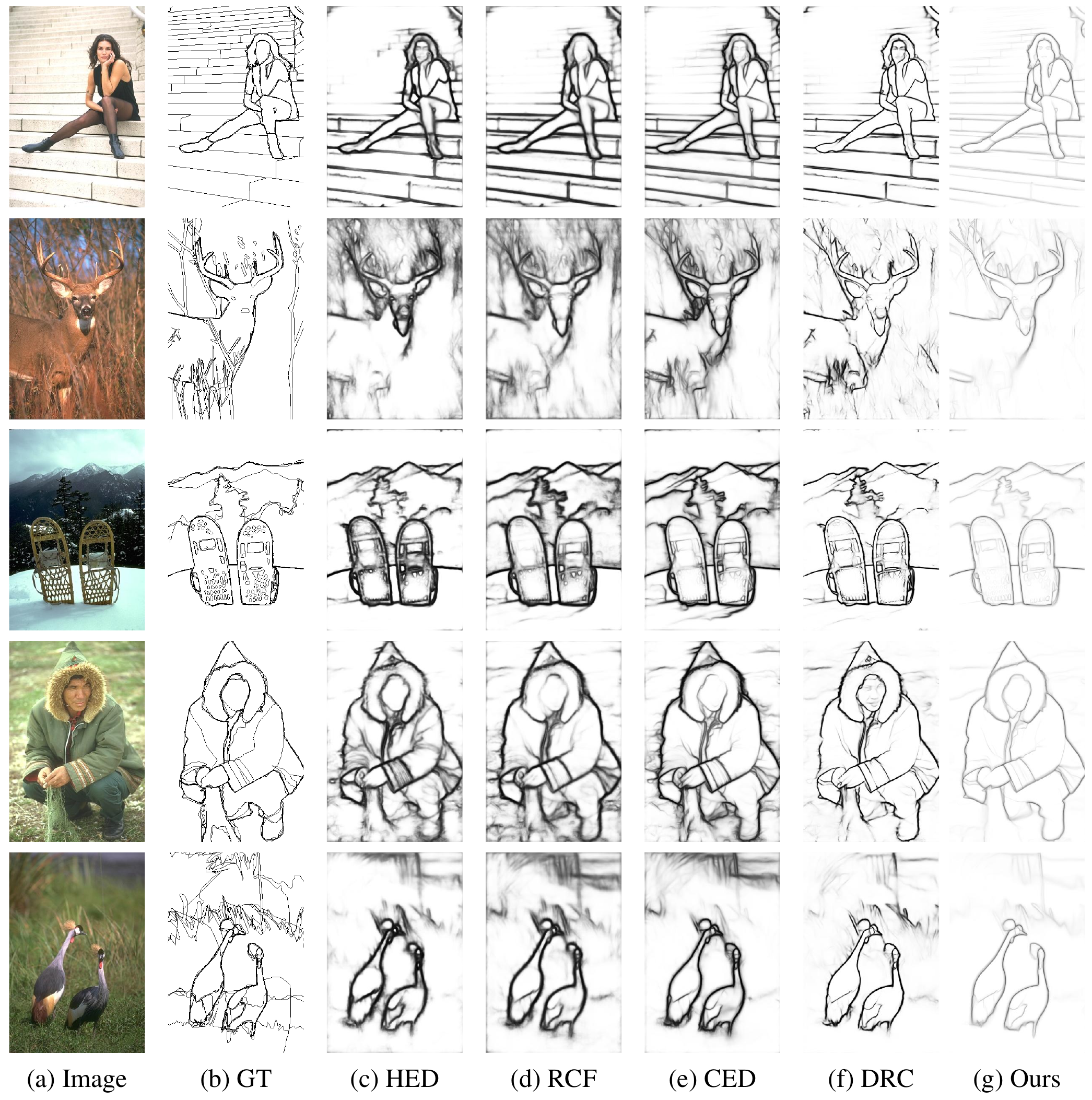}
	\caption{Some examples from SOTA methods on BSDS500.}
	\label{Fig.7}
\end{figure}

\begin{figure}[htbp]
	\centering
	\includegraphics[width=\textwidth]{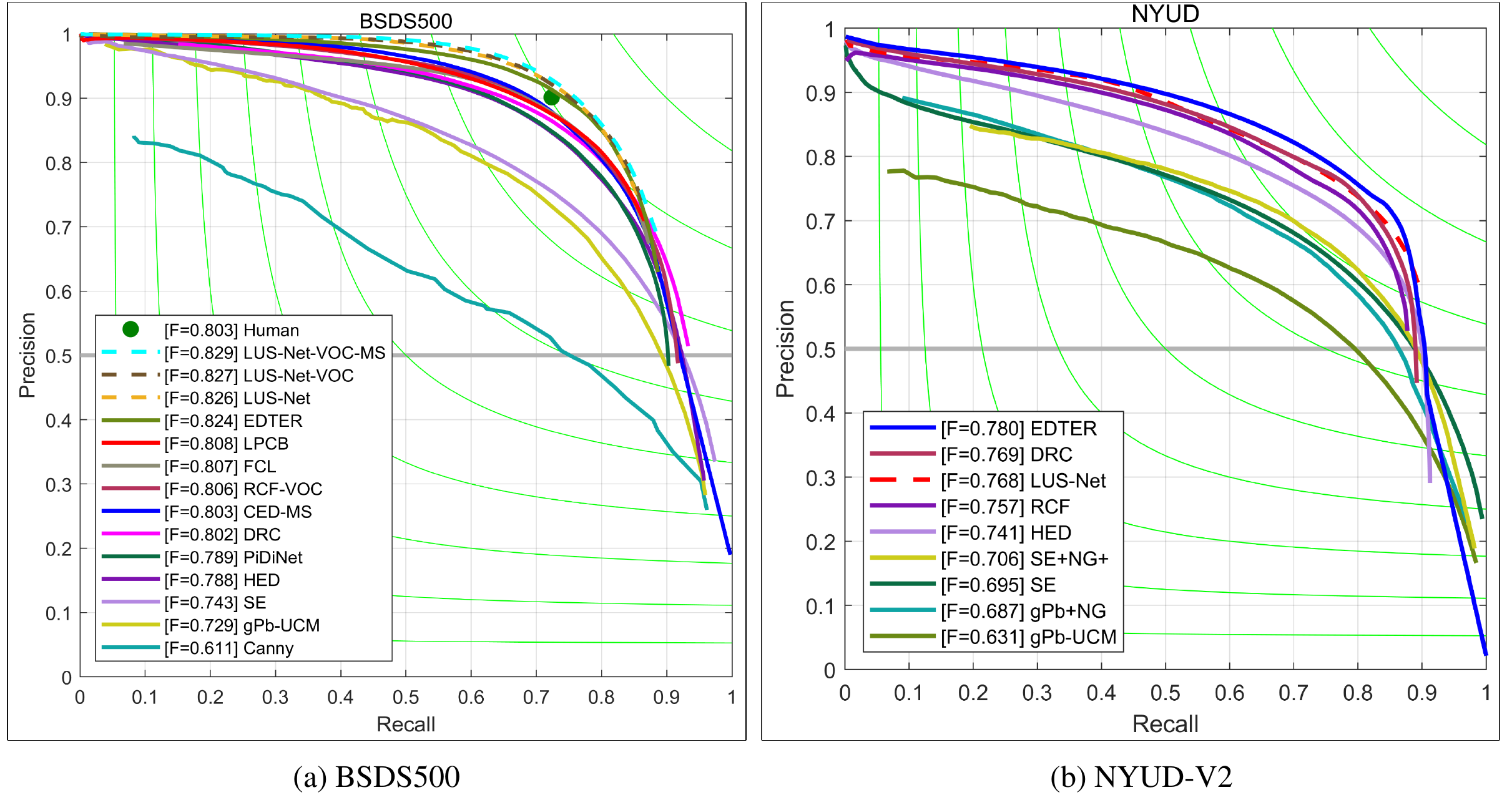}
	\caption{Precision-Recall curves on BSDS500 and NYUD-V2 dataset, respectively.}
	\label{PR_curves}
\end{figure}

\textbf{BSDS500:} Firstly, we compare our method with top-performing algorithms on BSDS500. We select recent state-of-the-art edge detectors which can be divided into two categories: the first category is methods without deep learning, which includes Canny \cite{canny1986computational}, gPb-UCM \cite{arbelaez2011contour}, and SE \cite{dollar2014fast};
the second category is approaches using deep learning technique, which including HED \cite{xie2015holistically}, RCF \cite{liu2017richer}, BDCN \cite{he2019bi}, CED \cite{wang2018deep}, LPCB \cite{deng2018learning}, DRC \cite{cao2020learning}, PiDiNet \cite{su2021pixel}, CATS \cite{huan2021unmixing}, FCL-Net \cite{xuan2022fcl}, EDTER \cite{pu2022edter}, Boosting-ED \cite{yang2023use}, DexiNed \cite{soria2023dense}, BetaNet \cite{li2023beta}, GCB-Net \cite{liu2024generating}, EdgeSAM \cite{yang2024boosting}, MuGE \cite{zhou2024muge}, and RankED \cite{cetinkaya2024ranked}. We additionally reference the studies of prior researchers who utilize extra training data sourced from the PASCAL VOC Context dataset \cite{mottaghi2014role} and adopt multi-scale testing to further improve the performance. Table \ref{Tab4} presents all metrics including S-Eval, C-Eval, Params, and FPS. Fig. \ref{Fig.7} shows some qualitative comparison results, and Fig. \ref{PR_curves} draws the Precision-Recall curves in S-Eval.

As shown in Fig. \ref{Fig.7}, HED and RCF are two classical algorithms for edge detection using VGG-16 \cite{simonyan2014very} architecture, while their edge maps are noisy and blurred. CED and DRC share the same goal as ours, which aims to generate crisp edge maps without any post-processing. However, compared to the other two methods, our method can generate higher-quality and crisper edge maps. Specifically, in the third row of Fig. \ref{Fig.7}, the contours of the man's eyes and nose are false edge pixels because they are not labeled in the corresponding groundtruth image, but our method can filter out these pixels, unlike the DRC which incorrectly classifies them. Similar results can be observed in other examples.

\begin{table}[htbp]
	\caption{Quantitative comparison on BSDS500 dataset. $\dagger$ indicates training with extra PASCAL VOC Context data. $\ddagger$ indicates the multi-scale testing. -S, -M, and -E respectively refers to ShuffleNetV2, MobileNetV2, and EfficientNetV2. $\star$ denotes the cited GPU speed, and $*$ denotes the inference speed tested on A40 GPU.}
	\label{Tab4}
	\centering
        \begin{tabular}{c|ccc|ccc|c|c}
            \toprule
            \multirow{2}{*}{Methods}& \multicolumn{3}{c|}{S-Eval}& \multicolumn{3}{c|}{C-Eval}& \multirow{2}{*}{Params}& \multirow{2}{*}{FPS}\\
            \cline{2-7}
            &ODS& OIS& AP& ODS& OIS& AP& & \\
            \hline
            Canny \cite{canny1986computational}& 0.611& 0.676& 0.520& -& -& -& -& -\\
            gPb-UCM \cite{arbelaez2011contour}& 0.729& 0.755& 0.745& -& -& -& -& -\\
            SE \cite{dollar2014fast}& 0.743& 0.764& 0.800& -& -& -& -& -\\
            \hline
            HED \cite{xie2015holistically}& 0.788& 0.808& 0.840& 0.576& 0.591& -& 14.7M& 30$\star$\\
            RCF \cite{liu2017richer}& 0.798& 0.815& -& 0.585& 0.604& -& 14.8M& 30$\star$\\
            DRC \cite{cao2020learning}& 0.802& 0.818& 0.800& 0.697& 0.705& -& 23.5M& 30$*$\\
            CED \cite{wang2018deep}& 0.794& 0.811& -& 0.642& 0.656& -& 14.9M& -\\
            CED$\ddagger$ \cite{wang2018deep}& 0.803& 0.820& -& -& -& -& 14.9M& -\\
            LPCB \cite{deng2018learning}& 0.800& 0.816& -& 0.693& 0.700& -& -& -\\
            BDCN \cite{he2019bi}& 0.806& 0.826& 0.847& 0.636& 0.650& -& 16.3M& 44$\star$\\
            PiDiNet \cite{su2021pixel}& 0.789& 0.803& -& 0.578& 0.587& -& \textbf{0.71M}& 69$*$\\
            CATS \cite{huan2021unmixing}& 0.800& 0.816& -& 0.666& 0.676& -& -& -\\
            FCL-Net \cite{xuan2022fcl}& 0.807& 0.822& -& 0.652& 0.666& 0.570& 16.46M& 7.8$*$\\
            EDTER \cite{pu2022edter}& 0.824& 0.841& 0.880& 0.698& 0.706& -& 468.84M& 2.1$\star$\\
            Boosting-ED$\dagger$ \cite{yang2023use}& 0.837& 0.853& 0.890& -& -& -& 468.84M& 2.1$\star$\\
            DexiNed \cite{soria2023dense}& 0.729& 0.745& 0.583& -& -& -& 35.2M& -\\
            BetaNet \cite{li2023beta}& 0.803& 0.822& -& -& -& -& -& -\\
            GCB-Net \cite{liu2024generating}& 0.811& 0.828& 0.809& 0.678& 0.686& 0.731& 14.97M& 61$*$\\
            EdgeSAM \cite{yang2024boosting}& 0.838& 0.852& 0.893& -& -& -& -& -\\
            MuGE \cite{zhou2024muge}& \textbf{0.850}& \textbf{0.856}& \textbf{0.896}& 0.671& 0.678& 0.713& 93.9M& 28$*$\\
            RankED \cite{cetinkaya2024ranked}& 0.824& 0.840& 0.895& 0.611& 0.619& 0.653& 114.4M& -\\
            \hline
            LUS-Net-S& 0.790& 0.812& 0.800& 0.670& 0.677& 0.640& \multirow{3}{*}{2.77M}& \multirow{3}{*}{\textbf{81.17}$*$}\\
            LUS-Net-S$\dagger$& 0.804& 0.825& 0.818& 0.683& 0.691& 0.712& &\\
            LUS-Net-S$\dagger\ddagger$& 0.813& 0.833& 0.839& 0.672& 0.677& 0.732& &\\
            LUS-Net-M& 0.805& 0.827& 0.805& 0.698& 0.705& 0.754& \multirow{3}{*}{11.51M}& \multirow{3}{*}{76.14$*$}\\
            LUS-Net-M$\dagger$& 0.815& 0.835& 0.820& 0.705& 0.711& 0.762& &\\
            LUS-Net-M$\dagger\ddagger$& 0.821& 0.842& 0.840& 0.684& 0.690& 0.740& &\\
            LUS-Net-E& 0.826& 0.846& 0.820& \textbf{0.720}& \textbf{0.726}& \textbf{0.779}& \multirow{3}{*}{70.41M}& \multirow{3}{*}{43.47$*$}\\         
            LUS-Net-E$\dagger$& 0.827& 0.846& 0.828& 0.717& 0.723& 0.777& &\\
            LUS-Net-E$\dagger\ddagger$& 0.829& 0.851& 0.846& 0.694& 0.700& 0.753& &\\
            \bottomrule
    \end{tabular}
\end{table}

Table \ref{Tab4} clearly shows that our LUS-Net-S still achieves a competitive performance when trained with only the BSDS500 dataset, outperforming HED, RCF, and PiDiNet in both S-Eval and C-Eval. When using the MobileNetV2 as the encoder, there are improvements in S-Eval and C-Eval, with ODS F-scores improving from 0.790 to 0.805 and 0.670 to 0.698, respectively. The highest performance is achieved by LUS-Net-E with ODS=0.826, OIS=0.846, and AP=0.820 in S-Eval. Compared to both EDTER, the first Transformer-based edge detection method, and the recent SOTA method RankED, our method demonstrates superior performance while maintaining a more efficient architecture with significantly fewer parameters (70.41M vs. 468.84M for EDTER and 114.4M for RankED). In addition, when we mix the extra PASCAL VOC Context data into the BSDS500 dataset, we observe a significant improvement in the S-Eval metrics (ODS=0.829, OIS=0.851, and AP=0.846), achieving competitive performance among all the methods. In the crispness evaluation, compared to existing methods addressing the edge thickness issue, including CED, LPCB, DRC, and GCB-Net, our LUS-Net-E demonstrates a significant performance margin, with ODS, OIS, and AP values of 0.720, 0.726, and 0.779 respectively, achieving SOTA performance among all the methods. Specifically, our ODS and OIS are 12.1\% and 10.7\% higher than those of CED, 3.9\% and 3.7\% higher than those of LPCB, 3.3\% and 3.0\% higher than those of DRC, and 6.2\% and 5.8\% higher than those of GCB-Net. These quantitative results are consistent with the qualitative visualization. It is worth noting that although our approach employs a single-granularity architecture in contrast to MuGE's multi-granularity design, and thus shows marginally lower performance under S-Eval, it exhibits remarkably better performance in C-Eval (0.720 vs. 0.611). In terms of model efficiency, our approach demonstrates impressive performance in both parameter efficiency and inference speed. Notably, LUS-Net-S achieves competitive performance with only 2.77M parameters while maintaining a high inference speed of 81.17 FPS on A40 GPU, surpassing previous crisp edge detection methods such as DRC \cite{cao2020learning} (23.5M, 30 FPS) and GCB-Net \cite{liu2024generating} (14.97M, 61 FPS). The larger variant LUS-Net-M (11.51M parameters, 76.14 FPS) and LUS-Net-E (70.41M parameters, 43.47 FPS) also maintain efficient parameter-performance trade-offs. All these comparison results demonstrate that our method has excellent performance, we successfully address the issue of edge thickness while improving the overall performance. Fig. \ref{PR_curves} shows that the performance of the human eye in edge detection is 0.803. Our results surpass the human level and achieve the best performance among current SOTA methods.

\textbf{NYUD-V2:} Secondly, we select the NYUD-V2 dataset to conduct another set of comparison experiments. We adopt some methods as before, which consists of algorithms without using deep learning such as gPb-UCM \cite{arbelaez2011contour}, gPb+NG \cite{gupta2013perceptual}, SE \cite{dollar2014fast} and SE+NG+ \cite{gupta2014learning}, and recent top edge detectors based on deep learning such as HED \cite{xie2015holistically}, RCF \cite{liu2017richer}, BDCN \cite{he2019bi}, CATS \cite{huan2021unmixing}, DRC \cite{cao2020learning}, PiDiNet \cite{su2021pixel}, DexiNed \cite{soria2023dense}, LPCB \cite{deng2018learning}, BetaNet \cite{li2023beta}, GCB-Net \cite{liu2024generating}, EDTER \cite{pu2022edter}, Boosting-ED \cite{yang2023use}, RankED \cite{cetinkaya2024ranked}, and EdgeSAM \cite{yang2024boosting}.


\begin{figure}[htbp]
	\centering
	\includegraphics[width=\textwidth]{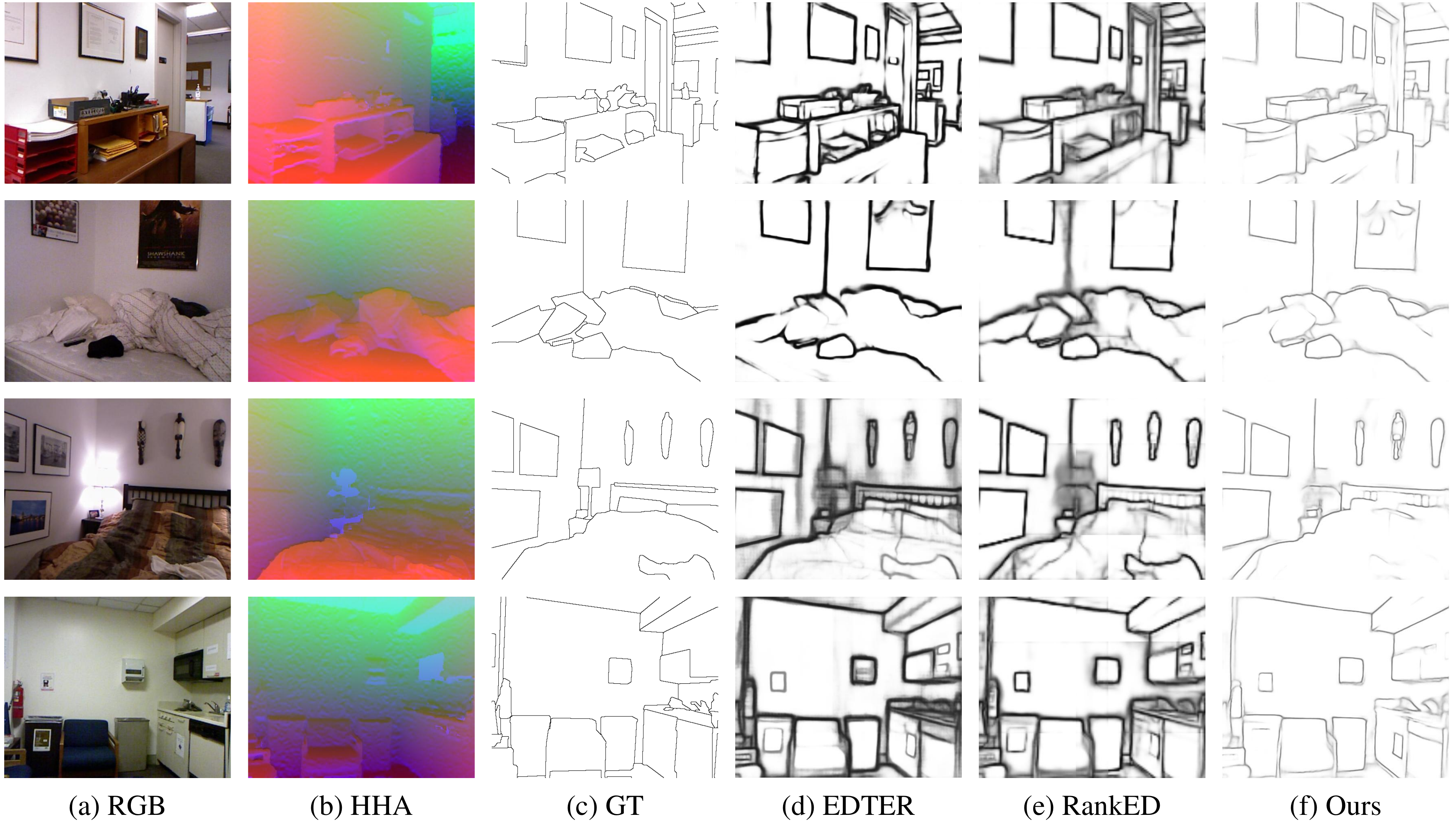}
	\caption{Some examples from SOTA methods on NYUD-V2.}
	\label{Fig.9}
\end{figure}

Since NYUD-V2 dataset consists of two types of images: RGB images and HHA images. We train and test our model on three versions following the previous researchers' works: (a) RGB images only; (b) HHA images only; (c) directly averaging the predictions from the RGB version and the HHA version. Some predicted examples are shown in Fig. \ref{Fig.9}. The quantitative evaluation results are summarized in Table \ref{tab_nyud} and the Precision-Recall curves are drawn in Fig. \ref{PR_curves}.

\begin{table}[htbp]
	\caption{Quantitative comparison results on NYUD-V2 dataset. $\dagger$ refers to training with extra BSDS500 data. $\star$ indicates the cited GPU speed, and $*$ denotes the inference speed tested on A40 GPU.}
	\label{tab_nyud}
	\centering
	\begin{tabular}{c|ccc|ccc|ccc|c|c}
	\toprule
	  \multirow{3}{*}{Methods}& \multicolumn{9}{c|}{S-Eval}& \multirow{3}{*}{Params}& \multirow{3}{*}{FPS}\\
        \cline{2-10}
        & \multicolumn{3}{c|}{RGB}& \multicolumn{3}{c|}{HHA}& \multicolumn{3}{c|}{RGB-HHA}& \\
        \cline{2-10}
        &ODS& OIS& AP& ODS& OIS& AP& ODS& OIS& AP& &\\
	  \hline
	  gPb-UCM \cite{arbelaez2011contour}& 0.631& 0.661& 0.562& -& -& -& -& -& -& -& \\
        gPb+NG \cite{gupta2013perceptual}& 0.687& 0.716& 0.629& -& -& -& -& -& -& -& -\\
	  SE \cite{dollar2014fast}& 0.695& 0.708& 0.719& -& -& -& -& -& -& -& -\\
        SE+NG+ \cite{gupta2014learning}& 0.706& 0.734& 0.549& -& -& -& -& -& -& -& -\\
	  \hline
	  HED \cite{xie2015holistically}& 0.720& 0.734& 0.734& 0.682& 0.695& 0.702& 0.746& 0.761& 0.786& 14.7M& 20$\star$\\
        RCF \cite{liu2017richer}& 0.729& 0.742& -& 0.705& 0.715& -& 0.757& 0.771& -& 14.8M& 20$\star$\\
        BDCN \cite{he2019bi}& 0.748& 0.763& 0.770& 0.707& 0.719& \textbf{0.731}& 0.765& 0.781& 0.813& 16.3M& 44$\star$\\
        CATS \cite{huan2021unmixing}& 0.732& 0.746& -& 0.693& 0.703& -& 0.755& 0.770& -& -& -\\
        DRC \cite{cao2020learning}& 0.749& 0.762& 0.718& 0.711& 0.722& 0.677& 0.769& 0.782& 0.771& 23.5M& 30$*$\\
        PiDiNet \cite{su2021pixel}& 0.733& 0.747& -& 0.715& 0.728& -& 0.756& 0.773& -& \textbf{0.71M}& \textbf{61$*$}\\
        DexiNed \cite{soria2023dense}& 0.658& 0.674& 0.556& -& -& -& -& -& -& 35.2M& -\\
        LPCB \cite{deng2018learning}& 0.739& 0.754& -& 0.707& 0.719& -& 0.762& 0.778& -& 15.7M& 30$\star$\\
        BetaNet \cite{li2023beta}& 0.734& 0.748& -& 0.714& 0.728& -& -& -& -&\\
        GCB-Net \cite{liu2024generating}& 0.751& 0.765& 0.719& 0.708& 0.720& 0.671& 0.766& 0.782& 0.777& 14.97M& 50$*$\\
        EDTER \cite{pu2022edter}& 0.774& 0.789& 0.797& 0.703& 0.718& 0.727& \textbf{0.780}& \textbf{0.797}& \textbf{0.814}& 468.84M& 2.1$\star$\\
        Boosting-ED$\dagger$ \cite{yang2023use}& 0.778& 0.793& 0.801& -& -& -& -& -& -& 468.84M& 2.1$\star$\\
        RankED \cite{cetinkaya2024ranked}& 0.780& 0.793& \textbf{0.826}& -& -& -& -& -& -& 114.4M& -\\
        EdgeSAM \cite{yang2024boosting}& \textbf{0.783}& \textbf{0.797}& 0.805& -& -& -& -& -& -& -& -\\
        \hline
        LUS-Net& 0.757& 0.768& 0.705& 0.717& 0.726& 0.644& 0.768& 0.780& 0.775& 70.41M& 34.26$*$\\
        \bottomrule
        \toprule
        \multirow{3}{*}{Methods}& \multicolumn{9}{c|}{C-Eval}& \multirow{3}{*}{Params}& \multirow{3}{*}{FPS}\\
        \cline{2-10}
        & \multicolumn{3}{c|}{RGB}& \multicolumn{3}{c|}{HHA}& \multicolumn{3}{c|}{RGB-HHA}& \\
        \cline{2-10}
        &ODS& OIS& AP& ODS& OIS& AP& ODS& OIS& AP& &\\
	  \hline
	  HED \cite{xie2015holistically}& 0.387& 0.404& -& 0.335& 0.350& -& 0.368& 0.384& -& 14.7M& 20$\star$\\
        RCF \cite{liu2017richer}& 0.395& 0.412& -& 0.333& 0.348& -& 0.374& 0.397& -& 14.8M& 20$\star$\\
        BDCN \cite{he2019bi}& 0.414& 0.439& -& 0.347& 0.367& -& 0.375& 0.392& -& 16.3M& 44$\star$\\
        CATS \cite{huan2021unmixing}& 0.405& 0.418& -& 0.349& 0.362& -& 0.382& 0.397& -& -& -\\
        PiDiNet \cite{su2021pixel}& 0.399& 0.424& 0.344& -& -& -& -& -& -& \textbf{0.71M}& \textbf{61$*$}\\
        DRC \cite{cao2020learning}& 0.403& 0.416& -& 0.370& 0.382& -& 0.403& 0.436& -& 23.5M& 30$*$\\
        GCB-Net \cite{liu2024generating}& 0.456& 0.468& 0.404& 0.391& 0.404& 0.337& 0.420& 0.433& 0.380& 14.97M& 50$*$\\
        EDTER \cite{pu2022edter}& 0.416& 0.438& 0.372& 0.394& 0.410& 0.348& 0.402& 0.420& 0.368& 468.84M& 2.1$\star$\\
        RankED \cite{cetinkaya2024ranked}& 0.477& 0.490& 0.474& -& -& -& -& -& -& 114.4M& -\\
        \hline
        LUS-Net& \textbf{0.540}& \textbf{0.554}& \textbf{0.508}& \textbf{0.435}& \textbf{0.450}& \textbf{0.400}& \textbf{0.513}& \textbf{0.524}& \textbf{0.501}& 70.41M& 34.26$*$\\
	    \bottomrule
	\end{tabular}
\end{table}

As shown in Fig. \ref{Fig.9}, (d), (e), and (f) are the output edge maps from EDTER, RankED, and our LUS-Net, respectively. The edge maps generated from our method exhibit consistent visualization with those of predictions on BSDS500. The predictions from our method are cleaner and crisper than the top edge detector EDTER and RankED, which demonstrates the effectiveness of our method.

Table \ref{tab_nyud} illustrates that our LUS-Net achieves strong performance across different image modalities (RGB, HHA, and RGB-HHA) and evaluation protocols (S-Eval and C-Eval). For RGB input under S-Eval, although our method achieves slightly lower scores (ODS=0.757) compared to recent methods like RankED (ODS=0.780) and EdgeSAM (ODS=0.783), it maintains competitive performance while using significantly fewer parameters (70.41M vs. 114.4M for RankED). More importantly, our method shows remarkable robustness in the challenging C-Eval protocol. For RGB input, LUS-Net achieves the ODS score of 0.540, significantly outperforming all recent SOTA methods including EDTER (0.416), RankED (0.477), and GCB-Net (0.456). This substantial performance gap (29.8\% higher than EDTER, 13.2\% higher than RankED, and 18.4\% higher than GCB-Net) demonstrates our method's enhanced capability in generating crisp edge maps. Similar robustness advantages can also be observed in HHA (0.435) and RGB-HHA (0.513) modalities, where our method consistently maintains superior performance across different image types. Specifically, LUS-Net achieves ODS=0.717 in HHA and ODS=0.768 in RGB-HHA under the S-Eval protocol, while maintaining strong robustness in C-Eval with ODS=0.435 in HHA and ODS=0.513 in RGB-HHA modalities. In terms of computational efficiency, LUS-Net achieves 34.26 FPS on an A40 GPU, which is significantly faster than transformer-based methods including EDTER (2.1 FPS) and Boosting-ED (2.1 FPS). The Precision-Recall curves in Fig. \ref{PR_curves} show that our LUS-Net has top performance. These results demonstrate that our method achieves an excellent balance between accuracy, robustness, and efficiency.

\textbf{BIPED:} Finally, we further evaluate our method on the BIPED dataset. Since the resolution of each image is relatively large ($1280\times720$), we crop each image-label pair randomly into $320\times320$ for training. We adopt recent SOTA methods for comparison which consist of HED \cite{xie2015holistically}, RCF \cite{liu2017richer}, BDCN \cite{he2019bi}, PiDiNet \cite{su2021pixel}, CATS \cite{huan2021unmixing}, FCL-Net \cite{xuan2022fcl}, DexiNed \cite{soria2023dense}, and EDTER \cite{pu2022edter}. 

\begin{figure}[htbp]
    \centering
    \includegraphics[width=\textwidth]{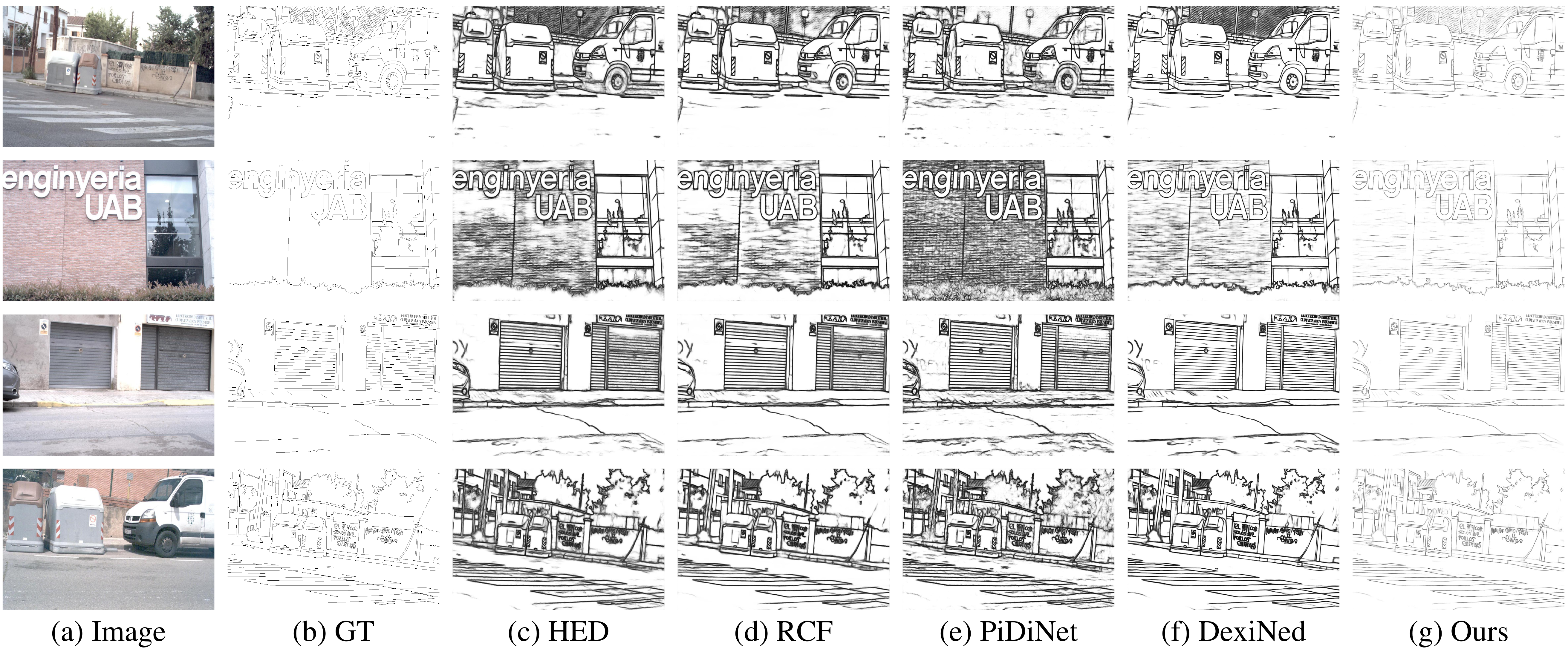}
    \caption{Some examples from SOTA methods on BIPED.}
    \label{BIPED_examples}
\end{figure}

As shown in Fig. \ref{BIPED_examples}, our LUS-Net demonstrates superior performance compared to existing SOTA approaches on the BIPED dataset. While HED and RCF demonstrate reasonable edge detection capabilities, they suffer from notable limitations including fragmented edges, inconsistent thickness, and discontinuity issues, particularly in complex architectural structures. PiDiNet occasionally misses fine structural details while DexiNed sometimes introduces spurious edges that deviate from the groundtruth. In contrast, our method demonstrates substantial advantages across multiple evaluation criteria, exhibiting enhanced edge continuity that maintains structural integrity while preserving fine-grained details often lost in competing approaches. Our method achieves superior noise suppression, producing cleaner edge maps with reduced artifacts, and demonstrates significantly improved edge localization accuracy with crisper and more precise boundary delineation that closely matches the groundtruth annotations.

The quantitative comparison results are listed in Table \ref{BIPED_Tab} and these results are similar to the performance shown on the BSDS500 and NYUD-V2. Specifically, our LUS-Net achieves ODS=0.902, OIS=0.908, and AP=0.912, surpassing most existing methods. When employing multi-scale testing, the performance further improves to ODS=0.903, OIS=0.909, and AP=0.930. Notably, LUS-Net is the first method to exceed 0.9, demonstrating its powerful performance. Furthermore, compared to other SOTA methods including FCL-Net, EDTER, and DexiNed, our approach achieves more efficient architectural design while maintaining superior performance. These results further validate the effectiveness and generalization ability of our method across different datasets and scenarios.

\begin{table}[htbp]
    \caption{Quantitative comparison results on BIPED dataset. $\ddagger$ refers to multi-scale testing.}
    \label{BIPED_Tab}
    \centering
    \begin{tabular}{c|ccc|c}
        \toprule
	  Methods& ODS& OIS& AP& Params\\
	  \hline
        HED \cite{xie2015holistically}& 0.829& 0.847& 0.869& 14.7M\\
	  RCF \cite{liu2017richer}& 0.849& 0.861& 0.906& 14.8M\\
        BDCN \cite{he2019bi}& 0.890& 0.899& \textbf{0.934}& 16.3M\\
        PiDiNet \cite{su2021pixel}& 0.868& 0.876& -& \textbf{0.71M}\\
        CATS \cite{huan2021unmixing}& 0.887& 0.892& 0.817& -\\
        FCL-Net \cite{xuan2022fcl}& 0.895& 0.900& -& 16.46M\\
        DexiNed \cite{soria2023dense}& 0.895& 0.900& 0.927& 35.2M\\
        EDTER \cite{pu2022edter}& 0.893& 0.898& -& 468.84M\\
	  \hline
        LUS-Net& 0.902& 0.908& 0.912& \multirow{2}{*}{70.41M}\\
        LUS-Net$\ddagger$& \textbf{0.903}& \textbf{0.909}& 0.930\\
	\bottomrule
    \end{tabular}
\end{table}

\newpage

\section{Conclusion}
\label{Discussion and conclusion}
\textbf{Discussion:} In this work, we proposed a simple yet effective U-shape network named LUS-Net for crisp edge detection. We leveraged the second-order derivative information to help the model locate true edge pixels more accurately. In addition, we constructed a novel hybrid focal loss function to solve the issue of imbalanced pixel distribution. Our method could generate crisp and clean contours without any post-processing, addressing the problem of edge thickness. The experiment results showed that we achieve state-of-the-art performance on three standard benchmarks, demonstrating the advantages and effectiveness of our method.

\textbf{Limitation:} There is still room for improvement. We currently rely on pre-trained backbones on ImageNet, which results in a cumbersome training process. Therefore, in future work, we will focus on exploring how to train the network from scratch, which could lead to further performance gains and a more efficient training pipeline.

\section*{Acknowledgments}
This work is supported by the National Key Research and Development Program of China (No.2022YFC3006302).

\bibliographystyle{unsrt}

\bibliography{cas-refs}


\bio[width=20mm,pos=l]{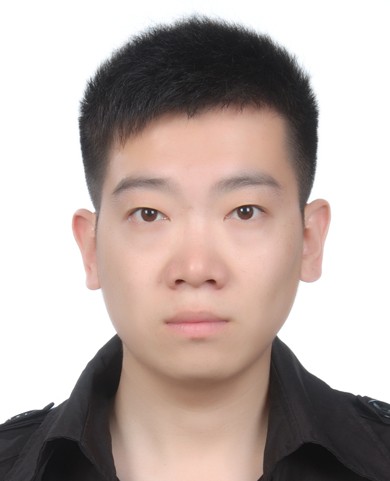}
\textbf{Changsong Liu} received the M.S. degree in Electronics and Communication Engineering from Tianjin International Engineering Institute, Tianjin University, China, in 2021. Currently, he is working toward the Ph.D. degree with the School of Microelectronics, Tianjin University. His areas of research are artificial intelligence, computer vision, and machine learning.
\endbio

\vspace{+10mm}

\bio[width=20mm,pos=l]{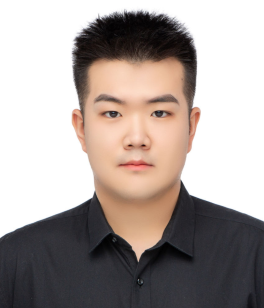}
\textbf{Yimeng Fan} received the B.S. degree in Electronic Science and Technology from Hebei University of Technology, China, in 2023. He is currently pursuing the M.S. degree in Electronic Science and Technology at Tianjin University, China. His research interests include deep learning and digital image processing.
\endbio

\vspace{+10mm}

\bio[width=20mm,pos=l]{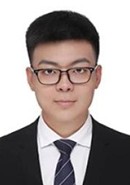}
\textbf{Mingyang Li} received the M.S. degree in Measuring Technology and Instrument from School of Precision Instrument and Opto-electronics Engineering, Tianjin University, China, in 2021. Currently, he is working toward the Ph.D degree with School of Microelectronics, Tianjin University. His areas of research are artificial intelligence, computer vision and machine learning.
\endbio

\vspace{+10mm}

\bio[width=20mm,pos=l]{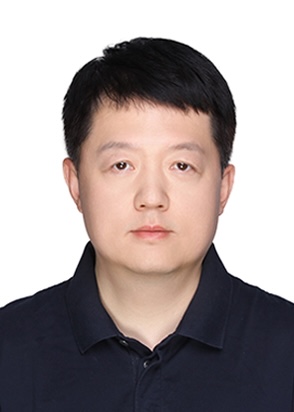}
\textbf{Wei Zhang} received the Ph.D. degree in Microelectronics and Solid-electronics from Tianjin University, China, in 2002. He is currently a Professor with School of Microelectronics, Tianjin University. His current research interests include artificial intelligence, data mining, data mining and digital image processing.
\endbio

\vspace{+10mm}

\bio[width=20mm,pos=l]{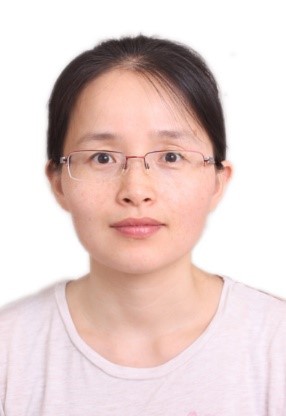}
\textbf{Yanyan Liu} received the B.S. and M.S. degrees in electrical engineering from Tianjin University, Tianjin, China, in 1999 and 2002, respectively, and the Ph.D. degree in electrical engineering from Nankai University, Tianjin, in 2010. In 2004, she joined the Optoelectronic Thin Film Device and Technology Research Institute, Nankai University, where she is currently an Associate Professor. Her current research interests include artificial intelligence and data mining.
\endbio

\vspace{+10mm}

\bio[width=20mm,pos=l]{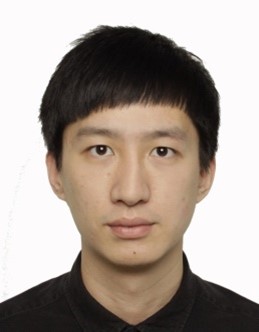}
\textbf{Yuming Li} received his M.S. degree in Electronic and Information Engineering from the Hong Kong Polytechnic University in 2017, and his Ph.D. degree in Microelectronics and Solid-electronics from the School of Microelectronics in Tianjin University, Tianjin, China, in 2023. He is now working as a Lecturer in the Luoyang Polytechnic, Luoyang, China. His research interests include image processing, pattern recognition and deep learning.
\endbio

\vspace{+10mm}

\bio[width=20mm,pos=l]{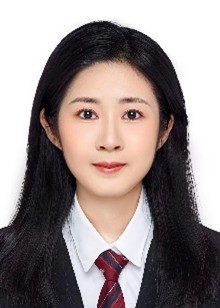}
\textbf{Wenlin Li} received the B.S. degree from School of Microelectronics, Tianjin University, in 2022. She is currently pursuing the M.S. degree with School of Microelectronics, Tianjin University. Her areas of research are object segmentation, computer vision and deep learning.
\endbio

\vspace{+10mm}

\bio[width=20mm,pos=l]{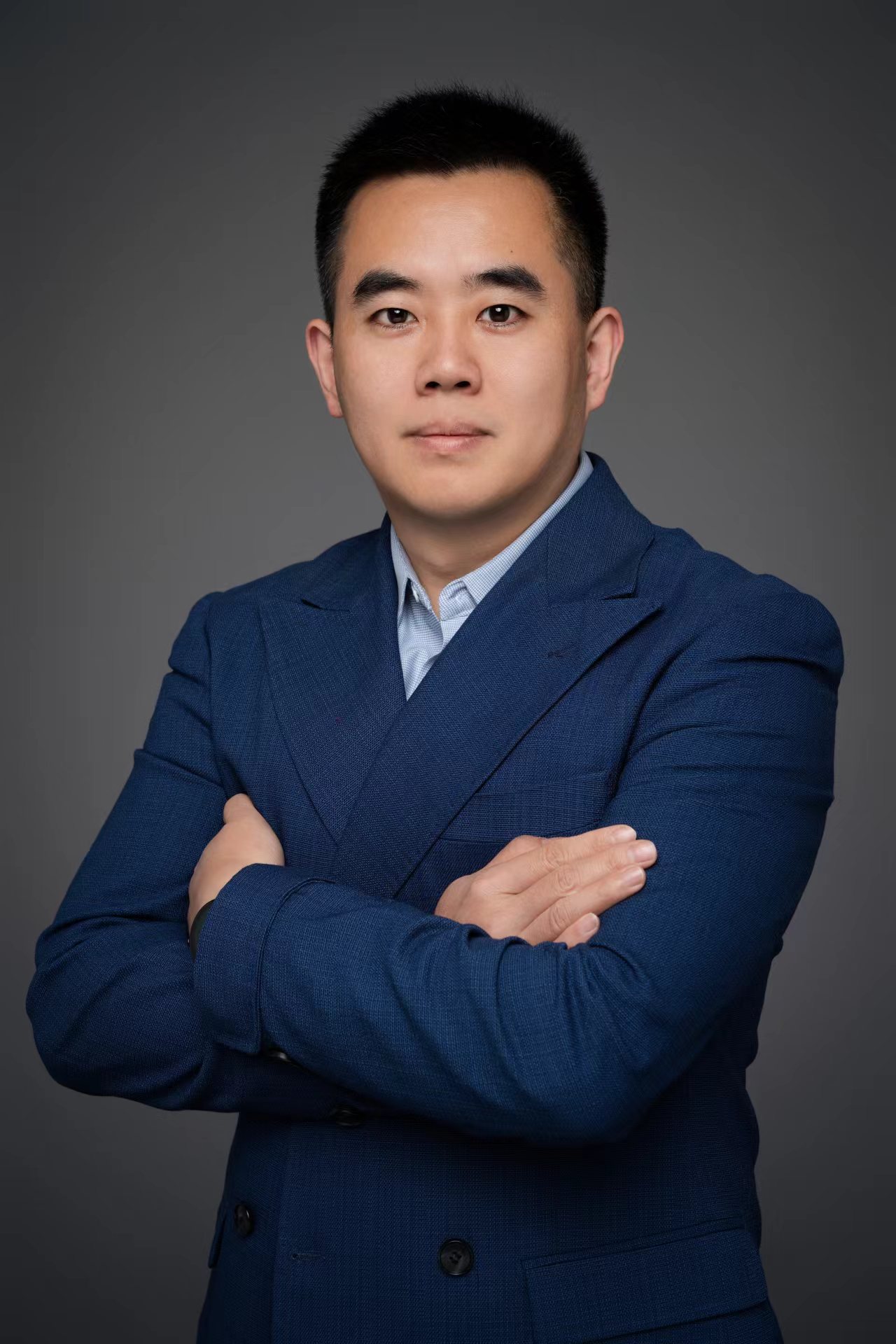}
\textbf{Liang Zhang} received the M.S. degree in Power Machinery and Engineering from Beihang University, China, in 2013. He works at Tianjin Fire Science and Technology Research Institute of MEM. His main research interests include fire investigation and automotive fire safety. He has been deeply involved in the hundreds of accident investigations of electric vehicles. Currently, he is working toward the Ph.D. degree at the School of Transportation Science and Engineering, Beihang University.
\endbio

\end{document}